\documentclass[smallcondensed]{svjour3}

\smartqed  

\usepackage{graphicx}
\usepackage[english]{babel}
\usepackage[utf8]{inputenc}

\usepackage{silence}
\usepackage{amsmath,amssymb}

\usepackage[linesnumbered]{algorithm2e}

\usepackage{multirow}

\usepackage{natbib}
\setcitestyle{open={(},close={)}}

\usepackage{url}
\usepackage{floatrow}

\usepackage{cases}

\usepackage{tcolorbox}
\usepackage{adjustbox}
\usepackage{graphicx,newfloat}
\DeclareFloatingEnvironment[
  fileext=lob,
  listname={List of Boxes},
  name=Box,
  placement=htp,
]{BOX}

\newcommand{\mat}[1]{\textbf{#1}}
\newcommand{\T}{^\top}
\newcommand{\inv}{^{-1}}
\newcommand{\defeq}{\equiv}

\renewcommand{\time}{t} 
\newcommand{\timeind}{k}

\newcommand{\E}{\mathbb{E}}
\newcommand{\M}{\mathcal{M}}
\newcommand{\N}{\mathcal{N}}
\renewcommand{\P}{\Xi}
\newcommand{\R}{\mathbb{R}}
\newcommand{\U}{\mathcal{U}}
\newcommand{\V}{\mathbb{V}}
\newcommand{\X}{\mathcal{X}}

\newcommand{\Z}{\mathcal{Z}}
\newcommand{\Zint}{\mathbb{Z}}
\newcommand{\GP}{\mathcal{GP}}

\newcommand{\BigO}[1]{\mathcal{O}(#1)}
\newcommand{\Tmeas}{{\mathcal{T}_\text{meas}}}

\newcommand{\thetamap}{\hat{\vec \theta}}

\renewcommand{\dim}{D}
\newcommand{\dimx}{{\dim_x}}
\newcommand{\dimp}{{\dim_{\theta}}}
\newcommand{\dimu}{{\dim_u}}
\newcommand{\dimxi}{{\dim_{x,i}}}
\newcommand{\dimpi}{{\dim_{\theta,i}}}

\newcommand{\dimz}{{\dim_z}}
\newcommand{\dimy}{{\dim_y}}

\newcommand{\xlow}{\underline{x}}
\newcommand{\xhigh}{\overline{x}}
\newcommand{\vxlow}{\underline{\vec x}}
\newcommand{\vxhigh}{\overline{\vec x}}

\newcommand{\xihigh}{\overline{\xi}}
\newcommand{\xibar}{\overline{\vec \xi}}

\DeclareMathOperator*{\argmax}{arg\,max}

\DeclareMathOperator*{\diag}{diag}

\DeclareMathOperator*{\cov}{cov}

\DeclareMathOperator*{\erf}{erf}

\renewcommand{\eqref}[1]{(\ref{#1})}
\newcommand{\figref}[1]{Fig.~\ref{#1}}
\newcommand{\boxref}[1]{Box~\ref{#1}}
\newcommand{\secref}[1]{Section~\ref{#1}}
\newcommand{\tabref}[1]{Table~\ref{#1}}
\newcommand{\appref}[1]{Appendix~\ref{#1}}

\newcommand\Tstrut{\rule{0pt}{2.4ex}}         % = `top' strut
\newcommand\Bstrut{\rule[-0.8ex]{0pt}{0pt}}   % = `bottom' strut

\usepackage{xcolor,colortbl}

\definecolor{columngray}{gray}{0.85}
\newcolumntype{a}{>{\columncolor{columngray}}c}

\usepackage{pifont}% http://ctan.org/pkg/pifont
\newcommand{\cmark}{{\color{green!50!black}\ding{51}}}%
\newif\ifreviewmode
\reviewmodetrue
\reviewmodefalse

\usepackage{color,soul}
\newcommand{\new}[1]{\hl{#1}}
\setstcolor{red}
\newcommand{\remove}[1]{\st{#1}}
\newcommand{\removecmd}[1]{#1}

\usepackage{todonotes}

\soulregister\cite7
\soulregister\citet7
\soulregister\citep7
\soulregister\citeauthor7
\soulregister\citeyear7
\soulregister\ref7
\soulregister\eq7
\soulregister\eqref7
\soulregister\appref7
\soulregister\figref7
\soulregister\secref7
\soulregister\boxref7
\soulregister\tabref7
\soulregister\pageref7
\soulregister\ul7
\soulregister\url7
\soulregister\remove7

\newcommand{\newmath}[1]{\colorbox{yellow}{$\displaystyle #1$}}
\newcommand{\rmmath}[1]{\colorbox{red}{$\displaystyle #1$}}

\usepackage{calc,xcolor}
\newcommand{\alignedbox}[1]{
  \begingroup\fcolorbox{yellow}{yellow}{$\displaystyle #1$} \endgroup
}
\usepackage{calc,xcolor}
\newcommand{\removealignedbox}[1]{
  \begingroup\fcolorbox{red}{red}{$\displaystyle #1$} \endgroup
}

\newcommand{\removesubsubsection}[1]{\subsubsection{\remove{#1}}}

\ifreviewmode
\else
    \renewcommand{\new}[1]{#1}
    \renewcommand{\remove}[1]{}
    \renewcommand{\alignedbox}[1]{#1}
    \renewcommand{\removealignedbox}[1]{}
    \renewcommand{\removecmd}[1]{}
    \renewcommand{\newmath}[1]{}
    \renewcommand{\rmmath}[1]{}
    \renewcommand{\removesubsubsection}[1]{} 
\fi

\begin{document}

% Title page and header information

% Title (Grants or other notes about the article that should go on the front page should
% be placed here. General acknowledgments should be placed at the end of the article)
%\title{Design of Dynamic Experiments for Black-Box Model Discrimination}
\title{\new{Using Gaussian Processes to Design Dynamic Experiments for Black-Box Model Discrimination under Uncertainty} %: Applications to Healthcare% using Gaussian Processes%
% \thanks{This work has received funding from the European Union’s Horizon 2020 research and innovation programme under the Marie Skłodowska-Curie grant agreement no. 675251, an EPSRC Research Fellowship (EP/P016871/1), and the Imperial Data Science Institute.}
}
% \titlerunning{Design of Dynamic Experiments for Black-Box Model Discrimination}  % if too long for running head
\titlerunning{Using GPs to Design Dynamic Experiments for Black-Box Model Discrimination}

%\author{Simon Olofsson \and \mbox{Eduardo dos Santos Schultz} \and Alexander Mitsos \and Marc Peter Deisenroth \and Ruth Misener}
\author{
Simon Olofsson\textsuperscript{1} 
\and 
\mbox{Eduardo S. Schultz}\textsuperscript{2} 
\and
Adel Mhamdi\textsuperscript{2}
\and 
Alexander Mitsos\textsuperscript{2} 
\and 
\mbox{Marc Peter Deisenroth}\textsuperscript{3} 
\and 
Ruth Misener\textsuperscript{1}
}
\authorrunning{Olofsson \emph{et al.}} % if too long for running head
%\authorrunning{Olofsson, Schultz, Mitsos, Deisenroth \& Misener} % if too long for running head

% "If available, the 16-digit ORCID of the author(s)" 
% Simon:     0000-0002-6997-9793
% Eduardo:   0000-0002-9038-8955
% Alexander: 0000-0003-0335-6566
% Marc:      0000-0003-1503-680X
% Ruth:      0000-0001-5612-5417

\institute{ 
    % \email{r.misener@imperial.ac.uk} 
    % \email{scw.olofsson@gmail.com} 
    \email{simon.olofsson15@alumni.imperial.ac.uk}
    \at
    \textsuperscript{1}\,%
    Department of Computing, Imperial College London, United Kingdom \\
    \textsuperscript{2}\,%
    Process Systems Engineering (AVT.SVT), RWTH Aachen, Germany \\
    \textsuperscript{3}\,%
    Department of Computer Science, University College London, United Kingdom 
    %
    %\and
    %ORCiD: \at
    %S.~Olofsson: 0000-0002-6997-9793 \\ 
    %E.~S.~Schultz: 0000-0002-9038-8955 \\
    %A.~Mitsos: 0000-0003-0335-6566 \\
    %M.~P.~Deisenroth: 0000-0003-1503-680X \\
    %R.~Misener: 0000-0001-5612-5417
}

\date{Received: date / Accepted: date}

\maketitle

\begin{abstract}
Diverse domains of science and engineering \remove{require and }use \new{parameterised }mechanistic \remove{mathematical }models\remove{, e.g.~systems of differential algebraic equations. Such models often contain uncertain parameters to be estimated from data}. \new{Engineers and scientists can often hypothesise several rival models to explain a specific process or phenomenon.} Consider a \remove{dynamic }model discrimination setting where we wish to \remove{chose: (i) what is}\new{find} the best mechanistic, \remove{time-varying}\new{dynamic} model \new{candidate }and \remove{(ii) what are }the best model parameter estimates. \remove{These tasks are often termed model discrimination/selection/validation/verification. }
Typically, several rival mechanistic models can explain \new{the available }data, so \remove{we incorporate available data and also run new experiments to gather more data. D}design of dynamic experiments for model discrimination helps optimally collect\new{ additional} data\new{ by finding experimental settings that maximise model prediction divergence}.
\remove{For rival mechanistic models where we have access to gradient information, we extend existing methods to incorporate a wider range of problem uncertainty and show that our proposed approach is equivalent to historical approaches when limiting the types of considered uncertainty. We also consider rival mechanistic models as dynamic \emph{black boxes} that we can evaluate, e.g.~by running legacy code, but where gradient or other advanced information is unavailable. We replace these black-box models with Gaussian process surrogate models and thereby extend the model discrimination setting to additionally incorporate rival black-box model. We also explore the consequences of using Gaussian process surrogates to approximate gradient-based methods.}
\new{We argue there are two main approaches in the literature for solving the optimal design problem: (i) the analytical approach, using linear and Gaussian approximations to find closed-form expressions for the design objective, and (ii) the data-driven approach, which often relies on computationally intensive Monte Carlo techniques. \citeauthor{Olofsson2018_ICML} (ICML 35, \citeyear{Olofsson2018_ICML}) introduced Gaussian process (GP) surrogate models to hybridise the analytical and data-driven approaches, which allowed for computationally efficient design of experiments for discriminating between black-box models. In this study, we demonstrate that we can extend existing methods for optimal design of \textit{dynamic} experiments to incorporate a wider range of problem uncertainty. We also extend the \cite{Olofsson2018_ICML} method of using GP surrogate models for discriminating between \textit{dynamic black-box} models.}
We \remove{also}\new{evaluate our approach on a well-known case study from literature, and} explore the consequences of using GP surrogates to approximate gradient-based methods. 
% \new{Our work shows one way machine learning can be used to hybridise--and bridge a gap between--existing methodologies, in this case approaches (i) and (ii) for design of experiments.}

\end{abstract}

\section{Introduction}

Mathematical models based on mechanistic assumptions are common in both engineering \new{and the natural sciences}, \remove{e.g.}\new{for example} pharmaceutical manufacturing (\remove{\citeauthor{RathoreWinkle2009}, \citeyear{RathoreWinkle2009}; }\citeauthor{Chen-etal:2019}, \citeyear{Chen-etal:2019}), \remove{and in the natural sciences, e.g.~}climate forecasting \new{\citep{Flato2014}} and high-energy physics (\citeauthor{Altarelli2014}, \citeyear{Altarelli2014}\remove{; \citeauthor{Mashnik2010}, \citeyear{Mashnik2010}; \citeauthor{Flato2014}, \citeyear{Flato2014}}). Researchers and engineers often hypothesise several such mechanistic models to explain the underlying system behaviour. Model discrimination differentiates between these rival hypotheses, i.e.~discards inaccurate models \citep{HunterReiner1965,BoxHill1967}. Often the available experimental data do not allow for discrimination between the rival models, \remove{i.e.}\new{when} all models sufficiently explain the available data\remove{, at least} within data accuracy\remove{, e.g.~\citet{mitsos_18_ident}} \new{\citep{mitsos_18_ident}}. This situation requires\remove{ more data, e.g.~by} running additional experiments \new{to collect more data}. But experiments are expensive, so we wish to discriminate between the models with as few additional experiments as possible
\remove{\citep{mitsos_17_oedminmax, mitsos_18_oedopd}}
% \new{\citep{mitsos_17_oedminmax, mitsos_18_oedopd}}.
% \new{\citep{mitsos_17_oedminmax}}.
\new{\citep{mitsos_18_oedopd}}.
%\todo[inline]{Here are some references (physics, physics, climate), which talk about model verification or testing: %\cite{Altarelli2014, Mashnik2010, Flato2014} % physics-physics-climate
%}

\remove{
As an example of requiring effective model discrimination, consider the parametric, mechanistic models used in the heavily regulated pharmaceutical manufacturing industry. The US Food \& Drug Administration (FDA) and European Medicines Agency (EMA) regulate the development, manufacturing and marketing of drugs and other pharmaceutical products. Drugs were not always heavily regulated: modern-day regulations 
%are the 
result from historical failures to protect patients
%The Elixir Sulfanilamide and thalidomide drug scandals are two of the most well-known examples 
\citep{Sulfanilamide}. 
%Thalidomide
Historical drug scandals led to improved regulations and thorough testing of new pharmaceutical products. 
}

\remove{
%
%Regulations at the beginning of the 21\textsuperscript{st} century were producing suboptimal results. Manufactured product quality was high, but pharmaceutical companies were reluctant to innovate because manufacturing process change required manufacturers to apply for new regulatory approval \citep{Yu2014}. 
%
To predict and prevent such tragedies while still encouraging innovation, both the FDA and EMA incorporate
\emph{quality-by-design} (QbD) 
%is a strategy that both  incorporate 
into their regulations
%: QbD both protects consumers and encourages innovation
\citep{RathoreWinkle2009}. QbD, which guarantees product quality through process design rather than by simply testing the final product, requires a deep understanding of a product and its manufacturing process. %The FDA guidelines on process validation \citep{FDAguidelines2011} now encourage manufacturers to collect and analyse data to identify opportunities for process improvements.
Typically, these QbD guidelines for pharmaceutical manufacturing require mechanistic understanding of the final product quality \citep{ICH_Q8}. These mechanistic models are typically parametric, with model parameters tuned to fit the experimental data, e.g.~as in \citet{lakerveld-etal:2013}.
}

\begin{figure}[!t]
	\centering
	\includegraphics[width=0.94\textwidth]{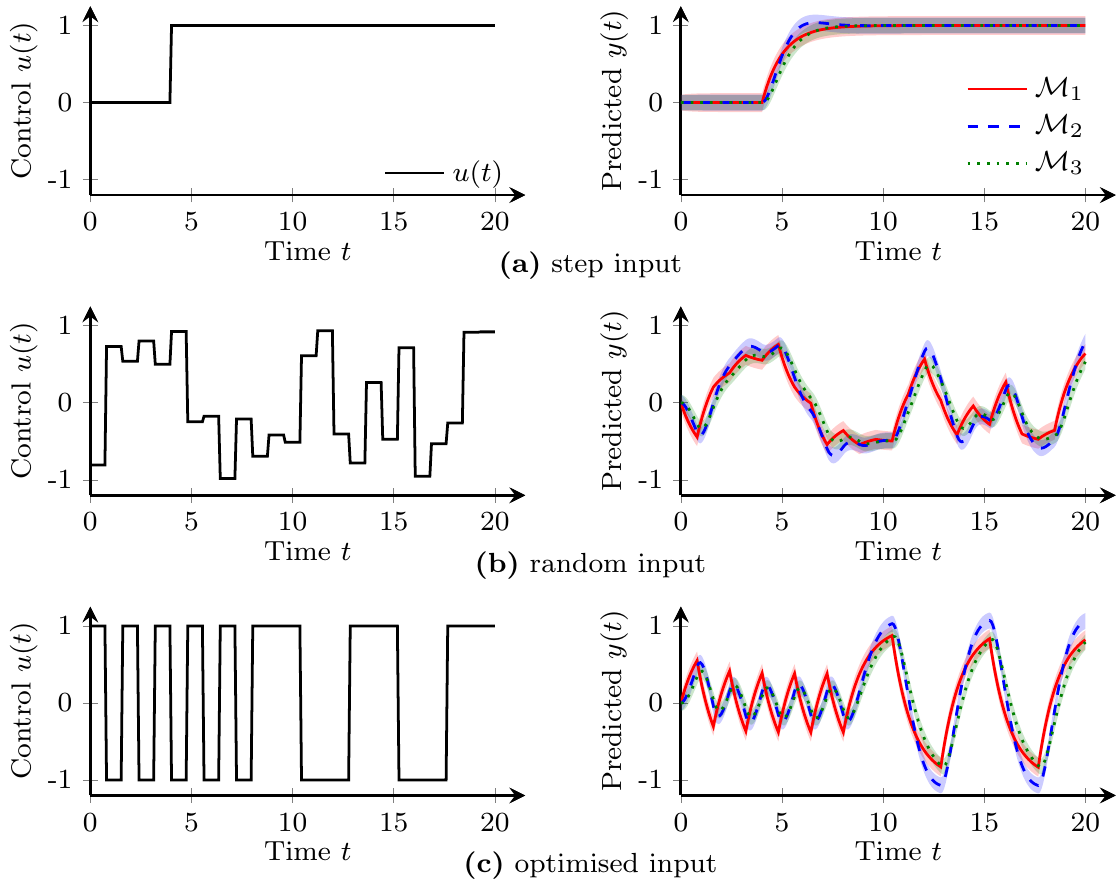}
	\vspace{-4mm}
	\caption{The three \cite{Bania2019} models $\M_1$, $\M_2$ and $\M_3$, defined in \remove{\eqref{eq:dode_example1}}\new{\appref{app:bania_models}}. \remove{(a), (c) and (e)}\new{The left-hand side} show three different piece-wise constant control input as function of time $\time$, and \remove{(b), (d) and (e)}\new{the right-hand side} show the corresponding model predictions\remove{, with} \new{(}mean and two standard deviations\new{)}.}
	\label{fig:dode_example}
\end{figure}

To illustrate model discrimination, consider three rival, linear continuous-time models $\M_1$, $\M_2$ and $\M_3$ \new{from }\remove{(}\citeauthor{Bania2019}\remove{,} \new{(}\citeyear{Bania2019})\new{.}
% To illustrate model discrimination, consider three rival, linear continuous-time models \new{$\M_1$, $\M_2$ and $\M_3$, of the form $\mathrm{d} x = \left( \mat A_i x(\time) + \mat B_i u(\time) \right) \mathrm{d}\time + \mat C_i \mathrm{d} w$}\remove{$\M_1$, $\M_2$ and $\M_3$} \citep{Bania2019}\new{,}
\setcounter{equation}{-1}
\removecmd{\begin{align}
    \removealignedbox{
    \label{eq:dode_example1}
	\M_i:\quad \left\lbrace
	\begin{aligned}
		\left. \frac{\mathrm{d} x}{\mathrm{d}\time} \right|_{\time}
		&= \mat A_i x(\time) + \mat B_i u(\time) + \mat C_i w (\time) \,,\\
		y_\timeind  &= [1,\,0,\,\dots,0] \, x(\time_\timeind) + 0.05 \cdot v_\timeind \,,
	\end{aligned}
	\right.
	}
\end{align}}
\remove{where $x(\time)$}\new{Let $y(\time)$} denote\remove{s} the \remove{state}\new{measurement} at time $\time$\remove{, $y_\timeind$ denotes measurements of a subset of the states at discrete time points $\time_\timeind$, $\timeind=1, \dots, T$,} \new{and }$u(\time)$ \remove{denotes }the control input\remove{, $w(\time)$ is the process noise and $v(\time)$ is the measurement noise}.
%\remove{$w(\time)$}\new{$w$} is the process noise \new{given by a Wiener process with variance 1.} \remove{and $v(\time)$ is the measurement noise.}%\new{Measurements $y_k$ at discrete time points $\time_\timeind$, $\timeind=1, \dots, T$ are given by $y_\timeind = [1,\,0,\,\dots,0] \, x(\time_\timeind) + 0.05 \cdot v_\timeind$, where $v_\timeind \sim \N(0, 1)$ is Gaussian-distributed noise.}
\new{The models account for both process noise and measurement noise.}
\figref{fig:dode_example} shows the model predictions
%
% FIGURE
%
for three different piece-wise constant control signals $u(\time) \in [-1,\,1]$\remove{, with 
%\new{initial state $\vec x_0^{(i)} = [0,\,\dots,\,0]\T$, and} 
matrices $\mat A_i$, $\mat B_i$ and $\mat C_i$ defined in \appref{app:bania_models}}. \remove{The initial state is $\vec x_0^{(i)} = [0,\,\dots,\,0]\T$, and the process noise $w(\time) \sim \N(0, 1)$ and the measurement noise $v_\timeind \sim \N(0, 1)$ are Gaussian distributed.}
\figref{fig:dode_example}a shows a step input, and \figref{fig:dode_example}\remove{c}\new{b} uniformly distributed random inputs\remove{. \figref{fig:dode_example}b and \figref{fig:dode_example}d show}\new{, with} the corresponding approximate marginal predictive distributions over time. The different models $\M_1$, $\M_2$ and $\M_3$ yield different interpretations of underlying system mechanisms. But it is non-trivial to find control inputs yielding sufficiently different predictions to allow for model discrimination.

Observe in \figref{fig:dode_example}\remove{b}\new{a} and \figref{fig:dode_example}\remove{d}\new{b} that na\remove{i}\new{\"i}vely applying a step or random control input may not produce data resolving the model discrimination problem. A na\remove{i}\new{\"i}ve control input may also result in violations of system safety constraints. \figref{fig:dode_example}\remove{e}\new{c} shows an example of a different control signal (developed using the methods presented in this paper). This control signal is optimised to yield sufficiently large model predictions differences\remove{ (shown in \figref{fig:dode_example}f)} for model discrimination. More than one experiment may be required to discriminate between models, and a systematic approach is necessary to minimise the number of experiments.

\new{Another example of a model discrimination scenario comes from \cite{EspieMacchietto1989} (details in \appref{app:yeast_fermentation_cs}). This industrially relevant case study considers four rival models for yeast fermentation, where the controls are the bioreactor inlet velocity and substrate concentration. Yeast fermentation is a very common process in pharmaceutical manufacturing, but the abundance of highly specialised yeast strains \citep{yeast_strains} make modelling and optimising the processes more difficult.}

The example\new{s} provided by \cite{Bania2019}\new{ and \cite{EspieMacchietto1989}} give\remove{s} easy access to model gradients. But many industrially relevant models\new{, such as those implemented in legacy code,} cannot be written in closed form\remove{, e.g.~if they are implemented in legacy code}. From an optimisation perspective, these models are effectively black boxes. We can evaluate the models, but the gradients are not readily available. Specialist software can generate the gradients automatically, but this poses substantial requirements on the implementation of the specialist models \citep{naumann_book_1}. For model discrimination, we wish to be agnostic to the software implementation or model type, since this flexib\new{i}l\new{it}y \remove{(i) }allows for faster model prototyping and development, and \remove{(ii) }satisfies the personal preferences of researchers and engineers. \remove{This paper finds optimal control inputs and other experimental conditions, e.g.~the initial state $x(0)$, to discriminate dynamic white- or black-box models under uncertainty and subject to constraints.}

\new{In this study, we demonstrate that we can extend existing analytical methods for optimal design of dynamic experiments to incorporate a wider range of problem uncertainty. We also extend the \cite{Olofsson2018_ICML} method of using Gaussian process surrogate models, to the case of discriminating between \emph{dynamic} black-box models. We use the well-known case study from \cite{EspieMacchietto1989} to show that our proposed approach is equivalent to historical approaches when limiting the types of considered uncertainty.}

\section{Design of Dynamic Experiments for Model Discrimination}

We consider state-space formulations of dynamic models \citep[Ch.~3]{statespacebook}.
% Prior work considers either discrete- or continuous-time models.
Most of this paper focuses on discrete-time models, but Appendi\remove{x}\new{ces} \ref{app:cont_model}\new{ and \ref{app:disc_delta_model}, respectively,} additionally show\remove{s} how to extend our contributions to continuous-time models\new{ and discrete-time models with a $\Delta$-transition, which follows from an Euler discretisation of continuous-time dynamics (\citeauthor{eulermethodbook}, \citeyear{eulermethodbook}, Ch.~2)}.
% \appref{app:cont_model} and \appref{app:disc_delta_model}, respectively, extend our work to continuous-time models and discrete-time models with a $\Delta$-transition, which follows from an Euler discretisation of continuous-time dynamics \citep[Ch.~2]{eulermethodbook}.

Let $\mathcal{T} = \lbrace \time_0, \dots, \time_T \rbrace$ denote an ordered set of discrete time instances. These time instances are indexed by $\timeind = 0, \dots, T$, with $\timeind=0$ the starting time step of an experiment and $\timeind=T$ the final time step. For simplicity, we write $\timeind \in \mathcal{T}$. % in abuse of notation.
Assume the system at time step $\timeind$ is in a state $\vec x_\timeind = x(\time_\timeind) \in \R^\dimx$. The rate of change of the state is a function of the current state $\vec x_{\timeind}$, a control input $\vec u_{\timeind} = u(\time_{\timeind}) \in \R^\dimu$, and some process noise $\vec w_{\timeind}$. The models describing the rate of change are parameterised by $\vec \theta \in \R^\dimp$. The control signal $u(\time)$ is commonly discretised and piece-wise constant in order to make optimisation feasible in dynamic problems with continuous-time models\remove{, e.g.~in \cite{Schultz2019}}\new{ \citep{Schultz2019}}.
In general, we cannot observe all dimensions of the state $\vec x_{\timeind}$. This is commonly the case for pharmaceutical manufacturing where the models predict different chemical concentrations, only some of which can be measured. Let $\vec z_{\timeind} = z(\time_{\timeind}) \in \R^{\dimz}$ denote the observed state. We take measurements $\vec y_{\timeind} \in \R^\dimz$ of the observed states $\vec z_{\timeind}$ at discrete time instances $\timeind \in \Tmeas$. Measurement noise $\vec v_{\timeind}$ corrupts the measurements. % of the observed states.

\remove{\appref{app:cont_model} and \appref{app:disc_delta_model}, respectively, extend our work to continuous-time models and discrete-time models with a $\Delta$-transition, which follows from an Euler discretisation of continuous-time dynamics (\citeauthor{eulermethodbook}, \citeyear{eulermethodbook}, Ch.~2).}

%We may have continuous-time models, in which case the state $x(\time)$, the control input $u(\time)$, process noise $w(\time)$, and observed state $z(\time)$ are all functions of the time $\time$. For discrete-time models, we assume equidistant time points in $\mathcal{T}$, such that $\time_{\timeind+1} - \time_{\timeind} = \Delta \time$ for some $\Delta \time > 0$. Each time step $\timeind$ has a corresponding control input $\vec u_{\timeind}$ to the system, and the control signal $u(\time)$ is piece-wise constant, with $u(\time') \defeq \vec u_{\timeind}$ for $\time_{\timeind} \leq \time' < \time_{\timeind+1}$. Measurements $\vec y_{\timeind}$ are taken at times $\time_{\timeind} \in \Tmeas \subseteq \mathcal{T}$.

Let $\vec x_{0:T}$, $\vec z_{1:T}$ and $\vec u_{0:T-1}$ denote the sequences of states, observed states and control inputs, respectively. The inputs to the system have dimensionality $T \times \dimu + \dimp$, and the outputs have dimensionality $T \times \dimz$. %The control signal $u(\time)$ is commonly discretised and piece-wise constant in order to make optimisation feasible in dynamic problems with continuous-time models, e.g.\ in \cite{Schultz2019}.
Assume $M$ rival state space models $\M_1, \dots, \M_M$. Model $\M_i$ specifies a state transition, observation and measurement model given a state $\vec x_{\timeind}^{(i)}$, control input $\vec u_{\timeind}$ at time step $\timeind$, and model parameter vector $\vec \theta_i$
\setcounter{equation}{0}
\begin{alignat}{3}
    \label{eq:rival_models}
    \M_i:\quad
    \left\lbrace
    \begin{aligned}
        \vec x_{\timeind}^{(i)} &= 
        f_i(\vec x_{\timeind-1}^{(i)}, \,\vec u_{\timeind-1}, \,\vec \theta_i)
        + \vec w_{\timeind-1}^{(i)} \,, &&\text{(State transition)} \\
        \vec z_{\timeind}^{(i)} &= \mat H_i \vec x_{\timeind}^{(i)} \,, &&\text{(Observed states)} \\
        \vec y_{\timeind}^{(i)} &= \vec z_{\timeind}^{(i)} + \vec v_{\timeind}^{(i)} \,, &&\text{(Noisy measurement)} 
    \end{aligned}
    \right.
\end{alignat}
where $f_i$ is the transition function, and $\mat H_i$ is a matrix selecting the observed states. The number of states $\dimxi$ may differ between models $\M_i$, but the number of observed states $\dim_{z,i} = \dimz$ for all models. %The operator $\phi_i$ is defined as
%\begin{subnumcases}
%    {
%    \label{eq:phi_operator}
%    \phi_i \left[ \vec x_{\timeind-1}^{(i)}, \,\vec u_{\timeind-1}, \,\vec \theta_i \,\middle|\, f_i \right] 
%    =
%    }
%    %\begin{cases}
%        % Continuous-time model
%        \vec x_{\timeind-1}^{(i)}
%        + 
%        \int_{\time_{\timeind-1}}^{\time_{\timeind}} f_i (x^{(i)}(\time), \,\vec u_{\timeind-1}, \,\vec \theta_i ) \mathrm{d}\time \,,
%        %& \text{(a)} 
%        \label{eq:phi_operator_cont}
%        \\[2mm]
%        % Discrete-time model
%        f_i (\vec x_{\timeind-1}^{(i)}, \,\vec u_{\timeind-1}, \,\vec \theta_i ) \,,
%        %& \text{(b)} 
%        \label{eq:phi_operator_disc}
%        \\[2mm]
%        % Discrete-time models with a Delta-transition.
%        \vec x_{\timeind-1}^{(i)}
%        +
%        f_i (\vec x_{\timeind-1}^{(i)}, \,\vec u_{\timeind-1}, \,\vec \theta_i ) \,,
%        \label{eq:phi_operator_delta}
%        %& \text{(c)}
%    %\end{cases}
%\end{subnumcases}
%for continuous-time models in \eqref{eq:phi_operator_cont}, discrete-time models  in \eqref{eq:phi_operator_disc} and discrete-time models with a $\Delta$-transition  in \eqref{eq:phi_operator_delta}. The discrete-time model with a $\Delta$-transition follows from an Euler discretisation of continuous-time dynamics \citep[Ch.~2]{eulermethodbook}.
%, e.g.\ RESNET \citep{resnet}. 
%Note that the transition function $f$ would be different for the different models in \eqref{eq:phi_operator}.

We assume the observed state $\vec z_{\timeind}$ is a subset (or linear combination) \new{$\vec z_{\timeind} = \mat H \vec x_{\timeind}$ }of the state\remove{ $\vec x_{\timeind}$, i.e., $z(\time) = \mat H x(\time)$}. Since the transition function $f$ can be any nonlinear function, this is a minor restriction to the types of models we consider for domains such as pharmaceuticals and chemical manufacturing. Moreover, the methods described in this manuscript could be extended to the more general case of a nonlinear relationship $z(\time) = g(x(\time))$, where $g$ is known.

The fundamental principle of sequential experimental design for model discrimination says to select the next experimental point where the model predictions differ the most \citep{HunterReiner1965}. Due to the multiple sources of uncertainty\remove{, e.g.}\new{ (for example} measurement noise\remove{,}\new{)} this translates into maximising the divergence between the models' marginal predictive distributions. The problems are\new{, firstly,}\remove{ (i)} approximating the models' marginal predictive distributions, and\new{, secondly,}\remove{ (ii)} constructing and maximising a predictive distribution divergence measure. There are two main approaches to solve these problems: the analytical approach, and the data-driven approach.

The most common analytical approach assumes all uncertainty sources are Gaussian distributed and uses a first-order Taylor expansion around the input mean to propagate the Gaussian input distribution $p(\vec x_{\timeind-1}, \vec u_{\timeind-1}, \vec \theta)$ to a Gaussian output distribution $p(\vec x_{\timeind}\,|\,\dots)$ \citep{BoxHill1967, MacKay1992b, %AspreyMacchietto2000, 
ChenAsprey2003}. \appref{app:existing_work} offers a more detailed overview. Several closed-form divergence measures exist for Gaussian marginal predictive distributions \citep{BoxHill1967, BuzziFerraris1990, Michalik2010, Olofsson2019b}. The analytical approach may be computationally efficient, but \new{is also }limited in the sense that \remove{(i) }it \new{requires gradient information and }uses linear and Gaussian approximations to handle uncertainty\remove{, and (ii) it requires derivative information}.
%The measure of how much the models differ is the \textit{design criterion} and the design optimisation problem maximises this design criterion. There is a range of existing design criteria in literature that have closed-form expressions for rival models with Gaussian predictive distributions, and that are differentiable with respect to the mean and covariance of the input $\vec u_{1:T}$. Examples include the Mahalanobis distance between the means of the predictive distributions \citep{HunterReiner1965}, the Mahalanobis distance with parameter uncertainty \citep{BuzziFerraris1983}, the upper bound on change in Shannon entropy \citep{BoxHill1967, MacKay1992b}, the Akaike criterion weights \citep{Michalik2010}, and the quadratic Jensen-R\'enyi divergence \citep{Olofsson2019b}.

The data-driven approach uses Monte Carlo techniques to approximate the marginal predictive distributions' divergence
%\citep{Vanlier2014, Ryan2016, Woods2017}. 
\citep{Ryan2016}. 
Variational techniques can also be used \citep{Foster2019}. This approach is flexible, because it is agnostic to the model structure (and indifferent to the presence or absence of function gradients). It may also be more accurate given enough samples, since it does not rely on linearising the models or Gaussian distributed uncertainty. However, it is computationally expensive, does not scale well with the problem size and makes optimisation difficult.
%To the best extent of our knowledge, there is no existing work using a data-driven approach to tackle design of \textit{dynamic} experiments for model discrimination. 
As mentioned, the dimensionality of the input and output spaces for the rival models can grow large when we discretise in time. The computational cost associated with solving design of dynamic experiments problems using Monte Carlo techniques would likely become insurmountable.
\cite{Streif2014} and \cite{Paulson2019} use polynomial chaos methods for design of dynamic experiments, but must still rely on expensive approximations of the predictive distributions' divergence. \new{\appref{app:existing_work} offers a more detailed overview.}

\cite{Olofsson2018_ICML} propose a hybrid approach for black-box model discrimination. They replace the black-box models with Gaussian process (GP) surrogates.\new{ GPs are universal function approximators that allow for encoding prior knowledge about the underlying function. Unlike many other surrogate models, GPs also provide prediction uncertainty estimates.} Training data for the GP surrogates are collected by evaluating the original black-box model at sampled locations. The GP surrogates are then used in an analytical fashion. This approach is computationally relatively cheap and can accommodate black-box models. However, na\remove{i}\new{\"i}vely applying the \cite{Olofsson2018_ICML} approach to design of dynamic experiments is intractable, since it would operate directly on the mapping from $(\vec x_0, \vec u_{0:T-1}, \Tmeas)$ to $\vec z_{1:T}$. %input and output dimensionality both grow linearly with $T$. 
The GP surrogate input dimensionality then scales as $\BigO{\dimx + \dimu \times T + \dimp}$ and output dimensionality as $\BigO{\dimz \times T}$.
Even for small $T$, \remove{e.g.}\new{such as}~$T=20$, the dimensionality of the input space is too high for accurate GP inference. The number of GP surrogates would have to equal the output dimensionality, which would be very expensive memory-wise.

Columns (a)--(e) of \tabref{tab:related_work} summarise several proposed methods in design of dynamic experiments for model discrimination. %
\begin{table}[!t]
    \centering
    \begin{tabular}{l|c|c|c|c|c|a}
        Ref. & (a) & (b) & (c) & (d) & (e) & (f) \\
        \hline
        Nonlinear $f$                       & \cmark & \cmark &        & \cmark &(\cmark)& \cmark \Tstrut\Bstrut\\
        Discrete-time models                 &        &        & \cmark &        & \cmark & \cmark \Tstrut\Bstrut\\
        Continuous-time models               & \cmark & \cmark &        & \cmark &        & \cmark \Tstrut\Bstrut\\
        Black-box models                     &        &        &        & \cmark &        & \cmark \Tstrut\Bstrut\\
        \hline
        Measurement noise                    & \cmark & \cmark &(\cmark)& \cmark & \cmark & \cmark \Tstrut\Bstrut\\
        Process noise                        &        &        & \cmark &        & \cmark & \cmark \Tstrut\Bstrut\\
        Uncertain $\vec x_0$                 &        &        & \cmark & \cmark &        & \cmark \Tstrut\Bstrut\\
        Uncertain $\vec u_{\timeind}$                 &        &        &        &        &        & \cmark \Tstrut\Bstrut\\
        Uncertain $\vec \theta$              & \cmark &(\cmark)&        & \cmark &        & \cmark \Tstrut\Bstrut\\
        \hline
        Optimise $\vec x_0$                  & \cmark & \cmark &        &        &        & \cmark \Tstrut\Bstrut\\
        Optimise $\vec u_{0:T-1}$            & \cmark & \cmark & \cmark & \cmark & \cmark & \cmark \Tstrut\Bstrut\\
        Optimise $\mathcal{T}_\mathrm{meas}$ & \cmark & \cmark &        &        &        & \cmark \Tstrut\Bstrut\\
        Path constraints                     &        &(\cmark)&(\cmark)&        &        & \cmark \Tstrut\Bstrut\\
    \end{tabular}
    \caption{References: (a)~\cite{ChenAsprey2003}, (b)~\cite{SkandaLebiedz2013}, (c)~\cite{CheongManchester2014a}, (d)~\cite{Streif2014}, (e)~\cite{Bania2019}, (f) This work. Bracketed check marks: \cite{Bania2019} mention how their approach can be extended to non-linear transition functions $f$; \cite{CheongManchester2014a} has one noise signal that affects both states and measurements; \cite{SkandaLebiedz2013} use a robust problem formulation instead of marginalising out the model parameters; %\cite{ChenAsprey2003},
    \cite{SkandaLebiedz2013} and \cite{CheongManchester2014a} solve the design of experiments optimisation problem subject to path constraints that do not account for predicted state uncertainty.}
    \label{tab:related_work}
\end{table}%
%Therefore, we propose an extension to the \cite{Olofsson2018_ICML} GP approach, in order to solve design of \emph{dynamic} experiments for discrimination of black-box models. Column (f) shows the novelty of the approach proposed in this work compared to existing literature, e.g.\ accounting for more possible types of uncertainty as well as accommodating black-box models.
This paper extends both the analytical approach of \cite{ChenAsprey2003} and the \cite{Olofsson2018_ICML} GP approach, to design \emph{dynamic} experiments for discriminating black-box models. Column (f) shows the novelty of our proposed approach\remove{ compared to existing literature}, \remove{e.g.}\new{for example} analytically accommodating black-box models \remove{as well as}\new{and} accounting for \remove{more possible}\new{all the listed sources of }uncertainty\remove{ types}. The proposed approach's GP surrogate input dimensionality scales as $\BigO{\dimx + \dimu + \dimp}$ and output dimensionality as $\BigO{\dimx}$.

%
% GP regression
%
%\subsection{Gaussian Process Regression}

\subsection{Gaussian Process Regression}
\label{sec:gpregression}

%\new{We replace black-box models with Gaussian process surrogates.} \cite{Olofsson2018_ICML}

A Gaussian process (GP) is a collection of random variables, any finite subset of which is jointly Gaussian distributed. A GP is completely specified by a mean function $m$ and covariance function $k$ \citep{RasmussenWilliams2006}.
Assume observations $\vec q=[q_1,\dots,q_N]\T$ at locations $\mat R = [\vec r_1, \dots, \vec r_N]\T$, with $q_n \sim \N(g(\vec r_n), \sigma_\eta^2)$, of a latent function $g:\R^{\dim_r} \rightarrow \R$ and zero-mean i.i.d.\ Gaussian observation noise with variance $\sigma_\eta^2$. Let $\psi$ denote the GP hyperparameters, which consist of the observation noise variance $\sigma_\eta^2$ and covariance function $k$'s parameters. GP regression computes a predictive distribution $g(\vec r)\,|\,\mat R, \vec q, \psi \sim \N(\mu(\vec r_\ast), \sigma^2(\vec r_\ast))$ at an arbitrary test point $\vec r_\ast$, with
\begin{subequations}
\label{eq:gp_mean_and_var}
\begin{align}
    %\label{eq:gp_mean}
    \mu (\vec r_\ast) &= m(\vec r_\ast) + \vec k\T (\mat K + \sigma_\eta^2\mat I )\inv (\vec q - \vec m) \,, \\
    %\label{eq:gp_var}
    \sigma^2 (\vec r_\ast) &= k(\vec r_\ast, \vec r_\ast) - \vec k\T (\mat K + \sigma_\eta^2\mat I )\inv \vec k \,,
\end{align}
\end{subequations}
where $[\mat K]_{j,\ell} = k(\vec r_j, \vec r_\ell)$, $[\vec k]_j = k(\vec r_\ast, \vec r_j)$ and $[\vec m]_j = m(\vec r_j)$. The hyperparameters $\psi$ are typically learnt by maximising the marginal likelihood $p(\vec q\,|\, \mat R, \psi )$. % with respect to the hyperparameters.

For vector-valued functions\remove{, i.e.}\new{ (}vector fields\new{)}\remove{,} $g:\R^{\dim_r} \rightarrow \R^{\dim_q}$, we can place independent GP priors $\GP(m_{(d)}, k_{(d)})$ on each target dimension $d=1,\dots,\dim_q$, with corresponding covariance function hyperparameters $\psi_{(d)}$. This yields the posterior distribution $g(\vec r_\ast)\,|\,\mat R, \mat Q, \Psi \sim \newmath{\N}(M(\vec r_\ast), \Sigma(\vec r_\ast))$, with
\allowdisplaybreaks{
\begin{subequations}
\label{eq:vector_gp_mean_and_var}
\begin{align}
    M(\vec r_\ast) &= \big[\, \mu_{(1)}(\vec r_\ast), \dots, \mu_{(\dim_q)}(\vec r_\ast) \,\big]\T \,, \\
    \Sigma(\vec r_\ast) &= \diag\big(\,\sigma_{(1)}^2(\vec r_\ast), \dots, \sigma_{(\dim_q)}^2(\vec r_\ast)\,\big) \,,
\end{align}
\end{subequations}}%
where $\mu_{(d)}(\vec r_\ast)$ and $\sigma_{(d)}^2(\vec r_\ast)$ are the mean and variance given by \eqref{eq:gp_mean_and_var} of the $d^\mathrm{th}$ GP posterior, $\mat Q = [\vec q_{(1)}, \dots, \vec q_{(\dim_q)}]$, with $\vec q_{(d)}$ dimension $d$ of all observations, and $\Psi = \lbrace \psi_{(1)}, \dots, \psi_{(\dim_q)} \rbrace$ the joint set of hyperparameters. Independent GP priors can accommodate different covariance functions and hyperparameters for different target dimensions.

If the input \remove{is uncertain, e.g. }$\vec r_\ast \sim \N(\vec \mu_\ast, \vec \Sigma_\ast)$\new{ is uncertain}, the posterior distribution needs to account for the added uncertainty. The exact posterior distribution is generally intractable\remove{,} but can be approximated, \remove{e.g.~using}\new{for example with} moment matching
\begin{align}
    \label{eq:gp_uncertain_input}
    p \left( g(\vec r_\ast)\,|\,\mat R, \mat Q, \Psi, \vec \mu_\ast, \vec \Sigma_\ast \right)
    \approx \N \big(\,
    \E_{\vec r_\ast} \left[ M(\vec r_\ast) \right] ,\, 
    \E_{\vec r_\ast} \left[ \Sigma (\vec r_\ast) \right]
    +
    \V_{\vec r_\ast} \left[ M(\vec r_\ast) \right]
    \, \big) \,.
\end{align}
Computing the marginal mean and variance in \eqref{eq:gp_uncertain_input} is also often intractable, with the exception of some special cases \citep{QuinoneroCandela2003, deisenroth2009}. An alternative approximation uses a first-order Taylor expansion of the predictive mean and variance in \eqref{eq:gp_mean_and_var} with respect to $\vec r$, and propagates the input distribution through the linearisation
%\begin{align}
%\label{eq:mean_and_var_uncertain_input}
$    p \left( g(\vec r_\ast)\,|\,\mat R, \mat Q, \Psi, \vec \mu_\ast, \vec \Sigma_\ast \right)
    \approx \N \left(\,
    M(\vec \mu_\ast) ,\,
    \nabla_{\vec r} M \, \vec \Sigma_\ast \, \nabla_{\vec r} M \T
    +
    \Sigma(\vec \mu_\ast)
    \, \right)$
%\end{align}
where $\nabla_r M = \partial M(\vec r) / \partial \vec r |_{\vec r = \vec \mu_\ast}$.
The linearisation
%\eqref{eq:mean_and_var_uncertain_input} 
is analytically tractable. %as long as the derivatives $\nabla_r \mu$ are available. 
By construction, the covariance $\nabla_{\vec r} M \vec \Sigma_\ast \nabla_{\vec r} M\T + \Sigma(\vec \mu_\ast)$ is positive definite.

\section{Experimental Design for Model Discrimination Under Uncertainty}
\remove{We consider the general model formulation in \eqref{eq:rival_models}, which makes it possible to extend existing methods, particularly that of \cite{ChenAsprey2003}. We create a common optimal experimental design framework for discrimination between rival dynamic models with (i) analytical or black-box transition functions $f_i$, (ii) continuous- or discrete-time models, and (iii) multiple types of uncertainty.}
\new{We start from the state-space formulation in \eqref{eq:rival_models} to formulate the experimental design optimisation problem for discrimination between rival dynamic models. We aim to demonstrate a single experimental design framework for continuous- and discrete-time models, with analytical or black-box state transition functions. Firstly, we show how we can extend existing methods for design of experiments, particularly that of \cite{ChenAsprey2003}, to include more uncertainty terms, for example in the control input. Secondly, we show how the framework can accommodate GP surrogate models when the transition function is a black box. Thirdly, we outline some best practices (from literature and what we have found useful for GP surrogate models) for handling constraints in our optimisation problem.}

%
%
%
%\subsection{Problem Uncertainty}
In \eqref{eq:rival_models}, $\vec w_{\timeind}^{(i)}$ is zero-mean Gaussian-distributed process noise and $\vec v_{\timeind}^{(i)} \sim \N(\vec 0, \vec \Sigma_y)$ is independent and identically distributed measurement noise. The process noise $\vec w_{\timeind}^{(i)}$ has known covariance $\vec \Sigma_{x,i}$. % if $\M_i$ is a discrete-time model, or covariance $(\time_{\timeind+1}-\time_{\timeind}) \vec \Sigma_{x,i}$ if $\M_i$ is a continuous-time model. 
We assume the measurement noise covariance $\vec \Sigma_y$ is known but may be different ($\vec \Sigma_y = \vec \Sigma_y^{(i)}$) for different models $\M_i$\footnote{Note that model-specific measurement noise covariances interferes with some design criterion definitions.}. Apart from uncertainty due to process noise and measurement noise, we may have uncertainty in the control input, initial state and model parameters.

The control input $\vec u_{\timeind} \sim \N(\hat{\vec u}_{\timeind}, \vec \Sigma_{u,\timeind})$ at time step $\timeind$ is Gaussian distributed with mean given by a user-specified desired control input $\hat{\vec u}_{\timeind}$ and covariance $\vec \Sigma_{u,\timeind}$. The control covariance $\vec \Sigma_{u,\timeind}=\vec \Sigma_{u,\timeind}^{(i)}$ may be model-dependent. For simplicity, let the control inputs $\vec u_{\timeind}$ be piece-wise constant and the control covariance constant $\vec \Sigma_{u,\timeind}=\vec \Sigma_u$. Simple extensions of the framework could accommodate control inputs described e.g.\ by piece-wise polynomials or time-dependent control covariance. Let $\hat{\vec u}_{0:T-1} = \lbrace \hat{\vec u}_0, \dots, \hat{\vec u}_{T-1} \rbrace$ denote the sequence of user-specified control inputs.

The initial state $\vec x_0^{(i)} \sim \N(\vec \mu_0^{(i)}(\hat{\vec x}_0), \vec \Sigma_0^{(i)})$ depends on some user-specified initial state settings $\hat{\vec x}_0$ common for all models. The vector $\hat{\vec x}_0$ is the desired initial state. Model parameters $\vec \theta_i \sim \N(\thetamap_i, \vec \Sigma_{\theta,i})$ are Gaussian distributed with mean given by the maximum \emph{a posteriori} parameter estimate $\thetamap_i$. A Laplace approximation computes the model parameter covariance $\vec \Sigma_{\theta,i}$ \citep[Ch.~27]{MacKay2003}.

%
%
%
%\subsection{Problem Formulation}
The initial state settings $\hat{\vec x}_0$ and the control inputs $\hat{\vec u}_{0:T-1}$ determine the experimental outcomes. For continuous-time models $\M_i$, we may also want to optimise the measurement time points $\Tmeas$. Hence, we formulate the optimisation problem of design of dynamic experiments for model discrimination as
%\begin{small}
{\allowdisplaybreaks
\begin{subequations}
\label{eq:design_problem_formulation}
\begin{align}
    \begin{split}
    % Objective function
    %\argmax_{\hat{\vec x}_0,\, \hat{\vec u}_{0:T-1}, \Tmeas} \, 
    \argmax_{\substack{\hat{\vec x}_0,\, \hat{\vec u}_{0:T-1} \\ \Tmeas  }} \,
    & \sum_{\timeind \in \Tmeas} D \left(\vec y_{\timeind}^{(1)}, \dots, \vec y_{\timeind}^{(M)} \right)
    \end{split} 
    \\
    \begin{split}
    \mathrm{s.t.} \quad
    & \forall \timeind \in \lbrace 1, \dots, T \rbrace\,, \, \forall i \in \lbrace 1, \dots, M \rbrace: \\
    & \M_i:\,\,
    \left\lbrace
    \begin{aligned}
        \vec x_{\timeind}^{(i)} &= 
        %\phi_i( f_i, \vec x_{\timeind-1}^{(i)}, \vec u_{\timeind-1},\,\vec \theta_i) 
        %\phi_i \left[ \vec x_{\timeind-1}^{(i)}, \,\vec u_{\timeind-1}, \,\vec \theta_i \,\middle|\, f_i \right]
        f_i ( \vec x_{\timeind-1}^{(i)}, \vec u_{\timeind-1},\,\vec \theta_i) 
        + \vec w_{\timeind-1}^{(i)} \,, \\
        \vec z_{\timeind}^{(i)} &= \mat H_i \vec x_{\timeind}^{(i)} \,, \\ 
        \vec y_{\timeind}^{(i)} &= \vec z_{\timeind}^{(i)} + \vec v_{\timeind} \,, \\
    \end{aligned}
    \right.
    \end{split} 
    \\
    \begin{split}
    &
    \left.
    \begin{aligned}
        % Initial latent state bounds
        C_{x_0} \big( \hat{\vec x}_0 \big) &\geq \vec 0 \,,  \\
        % Control input bounds
        C_u \big( \hat{\vec u}_{\timeind} \big) &\geq \vec 0 \,, 
    \end{aligned}
    \,\,\,\,\,\middle|\,\,\,\,\,
    \begin{aligned}
        % Path constraint
        C_x \big( \vec x_{\timeind}^{(i)} \big) &\geq \vec 0 \,, \\
        % Path constraint
        C_z \big(\vec z_{\timeind}^{(i)} \big) &\geq \vec 0 \,,
    \end{aligned}
    \,\,\,\,\,\middle|\,\,\,\,\,
    \begin{aligned}
        % Time constraints
        C_\mathcal{T} \big( \Tmeas \big) &\geq \vec 0 \,. \\
        \hspace{1mm}
    \end{aligned}
    \right.
    \end{split}
\end{align}
\end{subequations}}%
%\end{small}
$D(\cdot)$ is the design criterion, i.e.\ a divergence measure between the predictive distributions. $C_{x_0}$, $C_u$, $C_x$, $C_z$ and $C_\mathcal{T}$ are constraints on the corresponding variables. %The operator $\phi_i$ is defined in \eqref{eq:phi_operator}. 

%The prediction of states is discussed in \secref{sec:latent_state_prediction}. The design criterion $D_{\ast\ast}$ is a divergence measure between the predictive distributions, discussed in \secref{sec:divergence_objective}. The constraints $C_u$, $C_z$ and $C_x$ are discussed in \secref{sec:constraints}.

%The optimisation problem in \eqref{eq:design_problem_formulation} is continuous and non-convex. If the number of measurements is restricted, such that $\mathcal{T}_\mathrm{meas} \subset \mathcal{T}$, the variable choice of $\mathcal{T}_\mathrm{meas}$ turns the optimisation problem into a mixed-integer non-convex problem. This new optimisation problem can be solved, but we will not consider it here.

%
% Latent state transition prediction
%
\subsection{State Transition}
\label{sec:latent_state_prediction}

Consider a single model $\M = \M_i$ with corresponding transition operator $\phi=\phi_i$. For a state distribution $\vec x_{\timeind} \sim \N(\vec \mu_{\timeind}, \vec \Sigma_{\timeind})$, the predicted observed state is $\vec z_{\timeind} \sim \N(\mat H \vec \mu_{\timeind},\,\mat H \vec \Sigma_{\timeind} \mat H\T)$ and the measurement distribution\remove{s} is $\vec y_{\timeind} \sim \N(\mat H \vec \mu_{\timeind},\,\mat H \vec \Sigma_{\timeind} \mat H\T + \vec \Sigma_y)$.
%{\allowdisplaybreaks
%\begin{align*}
%	\vec z_{\timeind} &\sim \N(\mat H \vec \mu_{\timeind},\,\mat H \vec \Sigma_{\timeind} \mat H\T) \,,\\
%	\vec y_{\timeind} &\sim \N(\mat H \vec \mu_{\timeind},\,\mat H \vec \Sigma_{\timeind} \mat H\T + \vec \Sigma_y) \,.
%\end{align*}}
Solving the optimisation problem in \eqref{eq:design_problem_formulation} requires the predictive distribution of the latent state $\vec x_{\timeind}$. This means propagating the uncertainty in the inputs to the transition function $f$ to its outputs. We assume we know \textit{a priori} whether $f$ is an analytical function or a black box, i.e.\ whether we \emph{do} (analytical) or \emph{do not} (black box) have derivative information of $f$ with respect to its inputs. The derivative information is required for closed-form uncertainty propagation from inputs to outputs using Taylor approximations. %, as in \secref{sec:classicalapproach} and \secref{sec:GP_surrogate_method}.

To obtain derivative information from a black-box transition function $f$, we place independent GP priors $f_{(d)} \sim \GP(m_{(d)}(\cdot),\, k_{x,(d)}(\cdot,\cdot) k_{u,(d)}(\cdot,\cdot) k_{\theta,(d)}(\cdot,\cdot))$ on output dimensions $d = 1, \dots, \dimx$ of $f$. %The motivation for using a GP surrogate for the transition function is discussed in \secref{sec:training_gps}. 
To simplify notation, let $\mu_f(\cdot)=\E_f[f(\cdot)]$ and $\Sigma_f(\cdot)=\V_f[f(\cdot)]$, such that
\begin{align}
	\label{eq:f_analytical_or_black_box}
    f(\cdot) \sim \N(\mu_f(\cdot),\Sigma_f(\cdot)) \defeq
    \left\lbrace
    \begin{aligned}
        &\N(f(\cdot),\vec 0) \,, &&\, f \text{ analytical} \\ %\text{(White-box)} \\
        &\N(\mu(\cdot),\Sigma(\cdot)) \,,\quad &&\, f \text{ black box} \\ %\text{(Black-box)}
    \end{aligned}
    \right.
\end{align}
where the posterior GP mean $\mu(\cdot)$ and covariance $\Sigma(\cdot)$ are given in \eqref{eq:vector_gp_mean_and_var}.

Given an initial state estimate $\vec x_0 \sim \N(\vec \mu_{0}, \vec \Sigma_{0})$, a sequence of control inputs $\vec u_{\timeind} \sim \N(\hat{\vec u}_{\timeind}, \vec \Sigma_{u})$, $\timeind=0,\dots,T-1$, a model parameter posterior $\vec \theta \sim \N(\thetamap, \vec \Sigma_\theta)$, and the state transition described by \eqref{eq:rival_models}, we wish to find the approximate state distribution $\vec x_{\timeind} \sim \N(\vec \mu_{\timeind}, \vec \Sigma_{\timeind})$ at any time step $\timeind \geq 1$, with mean and covariance given by the moments
\begin{subequations}
\label{eq:transit_state_mean_and_var}
\begin{align}
	\label{eq:transit_state_mean}
    \vec \mu_{\timeind} &= \E_{f, \vec x_0, \vec u_{0:t-1}, \vec \theta, \vec w_{0:t-1}} \left[ \vec x_{\timeind} \right] \,, \\
    \vec \Sigma_{\timeind} &= \V_{f, \vec x_0, \vec u_{0:t-1}, \vec \theta, \vec w_{0:t-1}} \left[ \vec x_{\timeind} \right] \,.
    \label{eq:transit_state_var}
\end{align}
\end{subequations}
Assume that the control covariance $\vec \Sigma_u$, model parameter covariance $\vec \Sigma_\theta$ and process noise covariance $\vec \Sigma_x$ are all constant and independent of $\vec x_{\timeind}$ and $\hat{\vec u}_{\timeind}$. %The latent state mean $\vec \mu_{\timeind}$ and covariance $\vec \Sigma_{\timeind}$ depend on the form of the transition operator $\phi$. %Three different types of transitions are considered here: discrete-time steps, discrete-time $\Delta$-transition steps, and continuous transitions.

Let $f(\tilde{\vec x}_{\timeind}) \sim \N(\mu_f(\tilde{\vec x}_{\timeind}), \Sigma_f(\tilde{\vec x}_{\timeind}))$ denote the transition function evaluated at the concatenated state, control input and model parameters $\tilde{\vec x}_{\timeind} = [\vec x_{\timeind}\T, \vec u_{\timeind}\T, \vec \theta\T]\T$, $ \tilde{\vec x}_{\timeind} \in \R^{\dimx + \dimu + \dimp}$. Assuming the state, control input and model parameters are Gaussian distributed, the concatenated vector has Gaussian distribution $\tilde{\vec x}_{\timeind} \sim \N(\tilde{\vec \mu}_{\timeind},\tilde{\vec \Sigma}_{\timeind})$ with
\begin{align}
	\label{eq:variable_dependence}
    \tilde{\vec \mu}_{\timeind} = \begin{bmatrix}
    		\vec \mu_{\timeind} \\ \hat{\vec u}_{\timeind} \\ \hat{\vec \theta}
    \end{bmatrix}
    \,, \quad
    \tilde{\vec \Sigma}_{\timeind} = \begin{bmatrix}
    		\vec \Sigma_{\timeind} & \vec 0 & \cov(\vec x_{\timeind},\,\vec \theta) \\
    		\vec 0 & \vec \Sigma_u & \vec 0 \\
    		\cov(\vec x_{\timeind},\,\vec \theta)\T & \vec 0 & \vec \Sigma_\theta
    \end{bmatrix} \,,
\end{align}
where $\cov(\vec x_0, \vec \theta) \defeq \vec 0$, and $\cov(\vec x_{\timeind},\, \vec u_{\timeind}) \defeq \vec 0$ (assuming $\vec u_{\timeind} \neq \vec u_{\timeind-1}$) since the latent state cannot depend on future control inputs. %The state transition $\vec x_{\timeind}, \vec u_{\timeind} \mapsto \vec x_{\timeind+1}$ in \eqref{eq:transit_state_mean_and_var} will be described in terms of updates on $\tilde{\vec \mu}_{\timeind}$ and $\tilde{\vec \Sigma}_{\timeind}$.

To simplify notation, let $\nabla_{\vec \xi} g$, with $g \in \lbrace f, \mu_f, \Sigma_f, \dots \rbrace$, denote the partial derivative of $g(\vec \xi)$ with respect to a variable $\vec \xi$, evaluated at the point $\E[\vec \xi]$.

%
% Discrete-Time Models
%
%Consider a discrete-time state space model, as in \eqref{eq:rival_models} with $\phi$ given by \eqref{eq:phi_operator_disc}. 
The discrete-time state space model assumes %the latent state transition is described by
%\begin{align*}
%    %\label{eq:dtss}
%    \M:\quad
%    \vec x_{\timeind+1}
%    = f ( \vec x_{\timeind}, \vec u_{\timeind}, \,\vec \theta ) + \vec w_{\timeind} \,,
%\end{align*}
%with 
process noise $\vec w_{\timeind} \sim \N(\vec 0, \vec \Sigma_x)$. Using a first-order Taylor expansion of $\mu_f(\tilde{\vec x})$ around $\tilde{\vec x}_{\timeind-1} = \tilde{\vec \mu}_{\timeind-1}$, the mean and variance of the latent state at time step $\timeind \geq 1$ in \eqref{eq:transit_state_mean} are approximately given by
\begin{align}
	\label{eq:transition_dtss}
	\begin{split}
    \vec \mu_{\timeind} &\approx \mu_f(\tilde{\vec \mu}_{\timeind-1}) \,,\\
	\vec \Sigma_{\timeind} 
	&\approx
	\nabla_{\tilde{\vec x}_{\timeind-1}} \vec \mu_{\timeind} 
	\tilde{\vec \Sigma}_{\timeind-1} 
	\left( \nabla_{\tilde{\vec x}_{\timeind-1}} \vec \mu_{\timeind} \right)\T
	+ 
	\vec \Sigma_x + \Sigma_f(\tilde{\vec \mu}_{\timeind}) \,, \\
	\cov(\vec x_{\timeind}, \vec \theta) 
	&\approx 
	\nabla_{\vec \theta} \vec \mu_{\timeind} \cov(\vec x_{\timeind-1}, \vec \theta)\T + \nabla_{\vec \theta} \vec \mu_{\timeind} \vec \Sigma_{\theta} \,.
	\end{split}
\end{align}
Note that $\nabla_{\tilde{\vec x}_{\timeind-1}} \vec \mu_{\timeind} \in \R^{\dimx \times (\dimx+\dimu+\dimp)}$. We calculate derivatives of $\vec \mu_{\timeind}$ and $\vec \Sigma_{\timeind}$ with respect to $\vec \mu_{\timeind-1}$, $\vec \Sigma_{\timeind-1}$ and $\hat{\vec u}_{\timeind-1}$.
%are calculated following the standard rules of matrix calculus. 
The derivatives require second-order derivative information of $f$ or the GP prediction. \appref{app:cont_model} and \ref{app:disc_delta_model} have similar expressions to \eqref{eq:transition_dtss} for the latent state transition for continuous-time models and discrete-time models with $\Delta$-transitions, respectively. \new{\appref{app:doe_algorithm} provides an algorithmic description of the design of experiments process.}

\subsubsection{Combining Original Transition Function and GP Surrogate}
\label{sec:exact_mean}
%The reason for replacing black-box transition functions $f$ with GP surrogates in \eqref{eq:f_analytical_or_black_box} is that GP surrogates allow for approximating the first- and second-order derivatives of $f$ with respect to its inputs without resorting to finite difference approximations. Finite difference approximations can be expensive if $f$ is expensive to evaluate or the input space is large, especially for second-order derivatives.
The transition function $f$ is replaced with a GP surrogate in \eqref{eq:f_analytical_or_black_box} to enable analytical approximations of first- and second-order derivatives of $f$.
Equation~\eqref{eq:f_analytical_or_black_box} presents a binary choice
%: either $f$ is analytical, in which case $\mu_f(\cdot) = f(\cdot)$ and $\nabla \mu_f(\cdot) = \nabla f(\cdot)$, or $f$ is a black box, in which case it is replaced entirely with a GP surrogate during prediction and $\mu_f(\cdot) = \mu(\cdot)$ and $\nabla \mu_f(\cdot) = \nabla \mu(\cdot)$. 
between an analytical approach where $\mu_f(\cdot) = f(\cdot)$, and a black-box approach where $\mu_f(\cdot) = \mu(\cdot)$.
%The idea is that the approximation $f(\cdot) \approx \mu(\cdot)$ is accurate enough that it allows us to design optimal experiments. However, the approximation may not be accurate enough for robust model discrimination.
%
There is a third possible approach, %besides the fully analytic and fully black-box approaches, 
where %$\mu_f(\cdot) = f(\cdot)$ and $\nabla \mu_f(\cdot) \approx \nabla \mu(\cdot)$, %both the original transition function $f$ and the GP surrogate are used. Consider the following approximations
%\begin{align*}
%	\E_{f, \vec x_{\timeind}, \vec u_{\timeind}, \vec \theta}[f(\vec x_{\timeind}, \vec u_{\timeind}, \vec \theta)] &\approx f(\vec \mu_{\timeind}, \hat{\vec u}_{\timeind}, \hat{\vec \theta}) \,, \\
%	\E_{f, \vec x_{\timeind}, \vec u_{\timeind}, \vec \theta}[\nabla f(\vec x_{\timeind}, \vec u_{\timeind}, \vec \theta)] &\approx \nabla \mu(\vec \mu_{\timeind}, \hat{\vec u}_{\timeind}, \hat{\vec \theta}) \,,
%\end{align*}
%i.e.\ use 
the black-box transition function $f$ computes the predictive mean, and the GP surrogate only approximates the derivatives of $f$. \tabref{tab:exact_transition_function} compactly shows the difference between the fully analytic approach (where $f$ is analytic), the fully black-box approach (where only the GP surrogate is used) and the proposed third approach (where both $f$ and the GP surrogate are used). The third approach limits the use of the GP surrogates to %the purpose for which they were introduced: 
approximating the gradients of $f$. 
%Note that the third approach does not yield the additional uncertainty term associated with the GP surrogates, since $\Sigma_f(\cdot) \defeq \vec 0$ when the exact transition function mean is used.
\begin{table}[!t]
	\centering
	\begin{tabular}{c | c | c}
		Fully analytic approach & Fully black-box approach & Third approach \Tstrut\Bstrut\\ \hline
		& & \\[-2mm]
		$\begin{aligned} 
			\mu_f(\cdot) &= f(\cdot) \,, \\
			\Sigma_f(\cdot) &= \vec 0 \,, \\
			\nabla \mu_f(\cdot) &= \nabla f(\cdot)
		\end{aligned}$ 
		& 
		$\begin{aligned} 
			\mu_f(\cdot) &= \mu(\cdot) \,, \\ 
			\Sigma_f(\cdot) &= \Sigma(\cdot) \,, \\
			\nabla \mu_f(\cdot) &= \nabla \mu(\cdot)
		\end{aligned}$ 
		&
		$\begin{aligned} 
			\mu_f(\cdot) &= f(\cdot) \,, \\ 
			\Sigma_f(\cdot) &= \vec 0 \,, \\
			\nabla \mu_f(\cdot) &\approx \nabla \mu(\cdot)
		\end{aligned}$ 
	\end{tabular}
	\caption{When $f$ is analytic we use the fully analytic approach in \eqref{eq:f_analytical_or_black_box}. When the transition function $f$ is a black box we may choose to replace it in our computations with a GP surrogate for a fully black-box approach. A third approach is to use the original black-box transition function $f$ for computing $\mu_f(\cdot)$ and the GP surrogate to approximate its gradients $\nabla \mu_f(\cdot)$.}
	\label{tab:exact_transition_function}
\end{table}

This third approach is appropriate to use during model discrimination, i.e.\ when analysing agreement between model predictions and experimental observations. %, even for expensive-to-evaluate transition functions during model discrimination. 
This reduces the risk that a model is discarded because of poor accuracy in the GP surrogate prediction.
%
%But there are also disadvantages associated with the third approach. 
If $f$ is expensive to evaluate, we may still choose to use the GP surrogate black-box approach to speed up \new{the }design of experiments\remove{, i.e.~when solving the optimisation problem} in \eqref{eq:design_problem_formulation}.
%
%We may also encounter numerical issues when the third approach is used when solving \eqref{eq:design_problem_formulation}. 
\tabref{tab:exact_transition_function} shows that for the fully analytic and black-box approaches, the gradient $\nabla \mu_f$ is exact, whereas for the third approach the gradient $\nabla \mu_f(\cdot) \approx \nabla \mu(\cdot)$ is an approximation. This may cause numerical issues when the GP mean $\mu(\cdot)$---or rather the gradient $\nabla \mu(\cdot)$---does not capture the behaviour in the transition function $f$ with sufficient accuracy. If a numerical solver is provided with inaccurate gradients it may\remove{ e.g.} converge slowly to a solution, time-out before reaching any solution, or even throw an error. %if a line search in the direction provided by a gradient fails to find a step in input space that will improve the value of the objective function. 
Therefore, from an optimisation point-of-view, it may be better to use the black-box approach than the third when solving the optimisation problem in \eqref{eq:design_problem_formulation}. %, even if the transition function $f$ is cheap to evaluate. 
On the one hand this means solving an approximation of the optimisation problem, %that we would ideally like to solve, 
but on the other hand we may be more likely to find a solution.

%
% Constraints
%
\subsection{Constraints}
\label{sec:constraints}

The optimisation problem in \eqref{eq:design_problem_formulation} is solved subject to constraints.
%on the initial state settings $\hat{\vec x}_0$, the sequence of control inputs $\hat{\vec u}_{0:T-1}$, the states $\vec x_{1:T}$, the observed states $\vec z_{1:T}$, and the measurement time points $\Tmeas$. 
%
%Constraints are common in optimisation problems due to physical or safety constraints in 
Constraints are commonly related to safety concerns or physical limitations of the real systems, \remove{e.g.}\new{for example} the maximum allowed electrical current in a machine or drug dose given to a patient.
However, we may also need to apply constraints on the states $\vec x_{1:T}$ if the transition function $f$ is replaced with a data-driven surrogate.
%, e.g.\ a GP surrogate (see \secref{sec:latent_state_constraints}).

The initial state $\hat{\vec x}_0$, the control inputs $\hat{\vec u}_{1:T}$ and the measurement time points $\Tmeas$ are independent, deterministic variables, set directly by the user. The states $\vec x_{1:T}$ and observed states $\vec z_{1:T}$ are dependent, stochastic variables. Constraints on independent and dependent variables are handled differently.

This section considers two types of constraints. Firstly, linear constraints
\begin{align}
	\label{eq:linear_constraint}
	\mat C_{\newmath{k}} \vec \xi_{\newmath{k}} - \xibar_{\newmath{k}} \geq \vec 0 \,,
\end{align}
where $\vec \xi_{\newmath{k}} \in \lbrace \hat{\vec x}_0, \hat{\vec u}_{\timeind}, \vec x_{\timeind}, \vec z_{\timeind}, \time_{\timeind} \rbrace$, $\vec \xi_{\newmath{k}} \in \R^{\dim_\xi}$ is an independent or a dependent variable, $\mat C_{\newmath{k}} \in \R^{\dim_C \times \dim_\xi}$ and $\xibar_{\newmath{k}} \in \R^{\dim_C}$, and the inequality is applied element-wise. \new{The constraint may also be time-independent, such that $\mat C_k = \mat C$ and $\vec \xibar_k = \vec \xibar$, for example for $\vec \xi = \hat{\vec x}_0$.} The second type of constraints are constraints on the absolute difference, \remove{e.g.}\new{for example}~the rate of change, between independent variables. We will not consider constraints on the absolute difference between stochastic, dependent variables.

\removesubsubsection{Independent Variable Constraints}
% \label{sec:indep_var_constraints}
\new{\textbf{Independent Variable Constraints}} 
The independent variables in the optimisation problem in \eqref{eq:design_problem_formulation} are the initial state settings $\hat{\vec x}_0$, the sequence of desired control inputs $\hat{\vec u}_{0:T-1}$ and the measurement time points $\Tmeas$ (for continuous-time models). \remove{Through Section 3.2.1, %\secref{sec:indep_var_constraints}, 
let $\vec \xi \in \lbrace \hat{\vec x}_0, \hat{\vec u}_{\timeind}, \time_{\timeind} \rbrace$ denote an independent variable.}
Linear constraints %such as in
%\eqref{eq:linear_constraint}-type linear constraints 
on independent, deterministic variables are straight-forward to handle. Note that the constraint in \eqref{eq:linear_constraint} is written in the format of constraints in the \eqref{eq:design_problem_formulation} optimisation problem.
% 
%Constraints on the absolute difference between independent variables are useful for two reasons: Firstly, there
There may be limitations (for physical or safety reasons) to how quickly the control input $\hat{\vec u}_{\timeind}$ can be varied.
%; Secondly, a minimum amount of time between measurements need to be enforced during optimisation. 
Let the absolute difference in dimension $d=1,\dots,\dimu$ of the control input between consecutive time steps be upper-bounded 
%by $\Delta_{u,(d)}$
as $| \hat{u}_{\timeind+1,(d)} - \hat{u}_{\timeind,(d)} | \leq \Delta_{u,(d)}$.
%\begin{align*}
%	\left| \hat{u}_{\timeind+1,(d)} - \hat{u}_{\timeind,(d)} \right| \leq \Delta_{u,(d)} \,.
%\end{align*}
Using a standard reformulation, this constraint can equivalently be written as
\begin{align*}
	\hat{u}_{\timeind+1,(d)} - \hat{u}_{\timeind,(d)} + \Delta_{u,(d)} \geq 0 
	\quad \wedge \quad 
	\hat{u}_{\timeind,(d)} - \hat{u}_{\timeind+1,(d)} + \Delta_{u,(d)} \geq 0 \,.
\end{align*}
%using the constraints format in \eqref{eq:design_problem_formulation}.

%
%
%
\removesubsubsection{Dependent Variable Constraints}
% \label{sec:dep_var_constraints}
\new{\textbf{Dependent Variable Constraints}} 
The dependent variables in the optimisation problem in \eqref{eq:design_problem_formulation} are the states $\vec x_{1:T}$ and observed states $\vec z_{1:T}$. Constraints on the dependent variables are typically more difficult to satisfy \citep{Fu2015}, because, as the name suggests, they are dependent on the initial state $\hat{\vec x}_0$ and the control sequence $\hat{\vec u}_{0:T-1}$. %Constraints on the dependent state variables $\vec x_{1:T}$ and $\vec z_{1:T}$ are often referred to as \textit{path constraints}. 

Let model $\M$ predict the state distribution $\vec x_{\timeind} \sim \N(\vec \mu_{\timeind}, \vec \Sigma_{\timeind})$ and observed state distribution $\vec z_{\timeind} \sim \N(\vec \mu_z, \vec \Sigma_z)$ at time step $\timeind$, where $\vec \mu_z = \mat H \vec \mu_{\timeind}$ and $\vec \Sigma_{z} = \mat H \vec \Sigma_{\timeind} \mat H\T$. \remove{Through Section 3.2.2, %\secref{sec:dep_var_constraints}, 
l}\new{L}et $\vec \xi_{\timeind} \sim \N(\vec \mu_\xi, \vec \Sigma_\xi) \in \lbrace \vec x_{\timeind}, \vec z_{\timeind} \rbrace$ denote a Gaussian-distributed dependent variable. \remove{We assume that the dependent variable constraint may be time-dependent, such that
%Expression 
\eqref{eq:linear_constraint} becomes}
\removecmd{\begin{align*}\removealignedbox{
	% \label{eq:path_constraint}
	\mat C_{\timeind} \vec \xi_{\timeind} - \xibar_{\timeind} \geq \vec 0 \,.
}\end{align*}}
To simplify notation, let $\P_{\timeind} \subset \R^{\dim_\xi}$ denote the space $\P_{\timeind} = \lbrace \vec \xi \,\,|\,\, \mat C_{\timeind} \vec \xi - \xibar_{\timeind} \geq 0 \mathrm{\,,} \,\rbrace$ at time step $\timeind$,
%\begin{align*}
%	%\label{eq:path_constraint_set}
%	\P_{\timeind} = \left\lbrace \vec \xi 
%	\,\,\middle|\,\, 
%	\mat C_{\timeind} \vec \xi - \xibar_{\timeind} \geq 0 \,,
%	\,\right\rbrace \,,
%\end{align*}
such that satisfying the linear constraint in \new{\eqref{eq:linear_constraint}} % \eqref{eq:path_constraint} 
at time $\timeind$ is equivalent to satisfying $\vec \xi_{\timeind} \in \P_{\timeind}$. Multiple sources of uncertainty affect the states $\vec \xi_{\timeind}$ and need to be accounted for. Constraints on dependent variables with unbounded probability distributions (\remove{e.g.}\new{such as} Gaussian distributions) are referred to as \emph{chance constraints}. The chance constraint equivalent of $\vec \xi_{\timeind} \in \P_{\timeind}$ is
$ %\begin{align}
%	\label{eq:path_chance_constraint}
P(\vec \xi_{\timeind} \in \P_{\timeind}) \geq 1 - \gamma %\,,
$, %\end{align}
for some $\gamma \in (0,1)$ that determines the chance constraint's confidence level requirement. Chance constraints 
%of the type in \eqref{eq:path_chance_constraint} 
are typically analytically intractable \citep[Ch.~11]{Prekopa1995}. The\remove{re is a} range of different tractable approximations for chance constraint\remove{, e.g.}\new{ include:}~the scenario approach \citep{CalafioreCampi2006}\remove{,}\new{;} the sample average approximation \citep{Pagnoncelli2009}\remove{,}\new{;} and the convex second-order cone approximation \citep{Mesbah2014b}. This paper only considers the cone approximation, and compares it to a dependent variable constraint that does not account for the uncertainty in the state $\vec \xi_{\timeind}$.

%
%
%
%\noindent
\emph{Mean Constraint} 
%\label{sec:mean_path_constraint}
We let the term \textit{mean constraint} denote the regular linear dependent variable constraint in \new{\eqref{eq:linear_constraint}} % \eqref{eq:path_constraint} 
operating on the state mean, i.e.\ requiring $\vec \mu_{\xi} \in \P_{\timeind}$. The mean constraint is used by existing literature on design of dynamic experiments (see \tabref{tab:related_work}). The mean constraint has the advantage of simplicity, at the expense of not accounting for the uncertainty in $\vec \xi_{\timeind}$, and is approximately equivalent to solving the optimisation problem subject to $P( \vec \xi_{\timeind} \in \P_{\timeind} ) \geq \left( \tfrac{1}{2} \right)^{\dim_\xi}$.
%\begin{align*}
%	P( \vec \xi_{\timeind} \in \P_{\timeind} ) \geq \tfrac{1}{2} \,,\quad \forall\, \timeind \in \mathcal{T} \,.
%\end{align*}
Hence, the mean constraint provides a poor guarantee that the constraint $\vec \xi_{\timeind} \in \P_{\timeind}$ will be satisfied for all time steps $\timeind$.

%
%
%
%\noindent
\emph{Cone Constraint} 
%\label{sec:box_path_constraint}
The convex second-order cone approximation %[\citeauthor{Mesbah2014b}, \citeyear{Mesbah2014b}] 
\citep{Mesbah2014b}
decomposes the linear dependent variable constraint in \new{\eqref{eq:linear_constraint}} % \eqref{eq:path_constraint} 
into multiple constraints
$ %$\begin{align}
	\vec c_{\timeind,(j)}\T \vec \xi_{\timeind} - \xihigh_{\timeind,(j)} \geq 0 \,,
$ %\end{align}
for $\mat C_{\timeind} = [\vec c_{\timeind,(1)}, \dots, \vec c_{\timeind,(\dim_C)}]\T$ and $\xibar_{\timeind} = [\xihigh_{\timeind,(1)}, \dots, \xihigh_{\timeind,(\dim_C)}]\T$. Each individual chance constraint $P(\vec c_{\timeind,(j)}\T \vec \xi_{\timeind} - \xihigh_{\timeind,(j)} \geq 0) \geq 1 - \gamma$ can be satisfied by
\begin{align*}
	\begin{bmatrix}
		\vec c_{\timeind,(j)}\T \\ \vec c_{\timeind,(j)}\T
	\end{bmatrix}
	\vec \mu_\xi
	+
	\alpha \sqrt{ \vec c_{\timeind,(j)}\T \vec \Sigma_\xi \vec c_{\timeind,(j)} }
	\begin{bmatrix}
		\phantom{-}1\, \\ -1\,
	\end{bmatrix}
	-
	\begin{bmatrix}
		\xihigh_{\timeind,(j)} \\ \xihigh_{\timeind,(j)}
	\end{bmatrix}
	\, \geq  \, \vec 0 \,,
\end{align*}
where $\alpha = \sqrt{2} \erf\inv(1-\gamma)$, with $\erf\inv(\cdot)$ the inverse error function. %\figref{fig:constraints}b illustrates the chance constraint for the example defined in \eqref{eq:mean_constraint_example}.

For a fixed covariance $\vec \Sigma_\xi$, the cone constraint is equivalent to the mean constraint for a smaller space $\hat{\P}_{\timeind} \subset \P_{\timeind}$. While mean constraint may poorly guarantee satisfying the chance constraint, the cone constraint may be overly conservative since it decomposes the full chance constraint into individual chance constraints.

\subsubsection{State Constraints for Data-Driven Surrogate Models}
\label{sec:latent_state_constraints}
%When using a GP surrogate for a black-box transition functions $f$, it is important to remember that GP regression is an interpolation---rather than an extrapolation---method, and may predict poorly (e.g.\ by reverting to its prior) in regions of the input space where there is little or no training data. 
When solving \eqref{eq:design_problem_formulation}, the predicted states $\vec \mu_{\timeind}$ may stray away from the state space region where there is state training data $\mat X = \lbrace \vec x_1, \dots, \vec x_N \rbrace$. This can cause numerical issues in the solver as the GP predictive variance grows large, and we may have reasons not to trust a corresponding allegedly optimal solution $\hat{\vec u}_{0:T-1}$. Hence $\vec \mu_{\timeind}$ should be appropriately constrained as $\vec \mu_{\timeind} \in \X$.

Assume we know the feasible control space $\U$ defined by the input constraints.
%The feasible\footnote{Feasible with regard to input constraints} control space $\U$ is assumed known, and 
Control input training data can be sampled appropriately to fill \remove{the control space}\new{$\U$}.
%We fix the model parameters, so it is sufficient to sample model parameter training data close to the maximum \emph{a posteriori} parameter estimate $\thetamap$. 
We sample model parameter training data in a small region around the maximum \emph{a posteriori} parameter estimate $\thetamap$\new{,}
%. This allows approximation of the gradient of $f$ with respect to $\theta$, e.g.\ required for the Laplace approximation of $\vec \Sigma_\theta$.
to \remove{compute estimated} gradient\remove{s}\new{ estimation} for the Laplace approximation of $\vec \Sigma_\theta$.
%\todo{to compute/estimate gradients for ...}
Assume the observed state $\vec z_{\timeind} = \mat H \vec x_{\timeind}$ is subject to a constraint $\vec z_{\timeind} \in \Z$, $\forall \timeind \in \mathcal{T}$. We w\remove{ould like}\new{ant} to sample state training data from a domain $\X^\ast = \lbrace \vec x_{\timeind} \,\,|\,\, \exists\,\vec u_{\timeind} \in \U \,:\, \mat H \vec x_{\timeind} \in \Z \iff \mat H \vec x_{\timeind+1} \in \Z \rbrace$, \remove{i.e.}\new{where} states $\vec x_{\timeind}$ \remove{(i) that }satisfy $\vec z_{\timeind} \in \Z$\remove{,} and \remove{(ii) for which }some control input $\vec u_{\timeind}$ generates an observed state $\vec z_{\timeind+1} \in \Z$.
Finding $\X^\ast$ is non-trivial even if the inverse transition function $f\inv$ is known. $\X^\ast$ can be approximated through exhaustive sampling, but this may be expensive.
Proceeding, we let $\Zint_x = \lbrace 1, \dots, \dimx \rbrace$ and assume $\X(\vxlow, \vxhigh) = \lbrace \vec x\,|\, \forall\, d \in \Zint_x \,:\, \xlow_{(d)} \leq x_{(d)} \leq \xhigh_{(d)} \rbrace \approx \X^\ast$ is a known hypercube.
%\begin{align*}
%	\X(\vxlow, \vxhigh) = \left\lbrace \vec x\,\,\middle|\,\, \forall\, d \in \Zint_x \,:\, \, \xlow_{(d)} \leq x_{(d)} \leq \xhigh_{(d)} \right\rbrace \,.
%\end{align*}
GP training data is sampled from $\X$. %In our implementation, 
We assume state training data is sampled from a grid (with samples on the borders of $\X$) and combined with %grid-sampled 
control training data and model parameter samples.\new{ }% drawn from a small uniform distribution centred around $\thetamap$.
To maintain a fixed training data density, the training data set has to grow exponentially with state dimensionality $\dimx$ and control input dimensionality $\dimu$.
%Additionally, 
%The noise variance hyperparameters $\sigma_{\eta,(d)}^2$ are assumed greater than zero for all $d = 1, \dots, \dimx$ independent GP priors' covariance functions.

For numerical stability when solving \eqref{eq:design_problem_formulation}, %especially for problems with initial states on the edge of $\X$, 
we propose %relaxing $\X$, by 
using state constraint 
%$\vec \mu_{\timeind} \in \tilde{\X}_{d_\text{out}}$ 
$\vec \mu_{\timeind} \in \X(\vxlow - \vec \chi_{d_\text{out}}, \vxhigh + \vec \chi_{d_\text{out}})$
for the GP surrogate corresponding to output dimension $d_\text{out}$ of $f$, $d_\text{out} \in \Zint_x$, %with
%\begin{align*}
%	\tilde{X}_{d_\text{out}} = \left\lbrace \vec x \,\,\middle|\,\, \forall\, d_\textbf{in} \in \Zint_x \rbrace \,:\, \, \xlow_{(d_\text{in})} - \chi_{d_\text{out},d_\text{in}} \leq x_{(d_\text{in})} \leq \xhigh_{(d_\text{in})} + \chi_{d_\text{out},d_\text{in}} \right\rbrace \,,
%\end{align*} 
where $\chi_{d_\text{out},(d)} \defeq |\xhigh_{(d)} - \xlow_{(d)}| / \lambda_{(d_\text{out}),(d)}$ and $\lambda_{(d_\text{out}),(d)}$ denotes input dimension $d$'s lengthscale hyperparameter of output dimension $d_\text{out}$'s GP surrogate's state covariance function $k_{x,(d)}$. This ensures $\X(\vxlow, \vxhigh) \subset \X(\vxlow - \vec \chi_{d_\text{out}}, \vxhigh + \vec \chi_{d_\text{out}})$, $\forall d_\text{out} \in \Zint_x$.
% \remove{This ensures $\X(\vxlow, \vxhigh) \subset \X(\vxlow - \vec \chi_{d_\text{out}}, \vxhigh + \vec \chi_{d_\text{out}})$,} \remove{$\forall d_\text{out} \in \Zint_x$.}

\section{Results}
\label{sec:results}

We present computational results for an \cite{EspieMacchietto1989} case study that considers yeast fermentation\remove{, with c}\new{. C}onstants\remove{, e.g.~for }\new{ (}true parameter values\remove{ and}\new{,} noise covariances,\new{ etc.) are} from \cite{ChenAsprey2003}.
Yeast fermentation is common in pharmaceutical manufacturing \citep{Martinez2012}. There are $\dimx=2$ states (biomass and substrate concentration) and $\dimu=2$ control inputs. We observe both states, hence $\dimz=\dimy=2$ and $\mat H_i = \mat H = \mat I$. The models have $\dimpi \in \lbrace 3, 4 \rbrace$ model parameters. \appref{app:yeast_fermentation_cs} describes the case study in detail.

\remove{We}\new{\secref{sec:results_comparison_to_literature}} present\new{s}\remove{ (i)} a comparison to literature results for the \cite{EspieMacchietto1989} case study, \remove{(ii)}\new{and \secref{sec:compare_path_constraints}} a performance comparison of mean and cone constraints\remove{, (iii)}\new{. In \secref{sec:control_uncertainty} we use} simulations \remove{with}\new{to assess the effect of} correctly modelled, underestimated or overestimated control signal uncertainty, and \remove{(iv) a}\new{in \secref{sec:gp_results} we} study \remove{of }the GP surrogate \new{model's }performance as analytic emulators of black-box transition functions.
%
% We generated all results in our open-source Python package for design of experiments, \textit{doepy}\footnote{\url{https://github.com/scwolof/doepy}}.
We \new{follow the design of experiments process outlined in Algorithm~\ref{alg:doe} in \appref{app:doe_algorithm}}\remove{generated all results using our open-source Python package for design of experiments, \textit{doepy}$^2$}. %\footnote{\url{https://github.com/scwolof/doepy}}.

%
% Comparison to Espie & Macchietto (1989) results
%
\subsection{Comparison to Literature}
\label{sec:results_comparison_to_literature}
\begin{figure}[!t]
	\centering
	\includegraphics[width=0.98\textwidth]{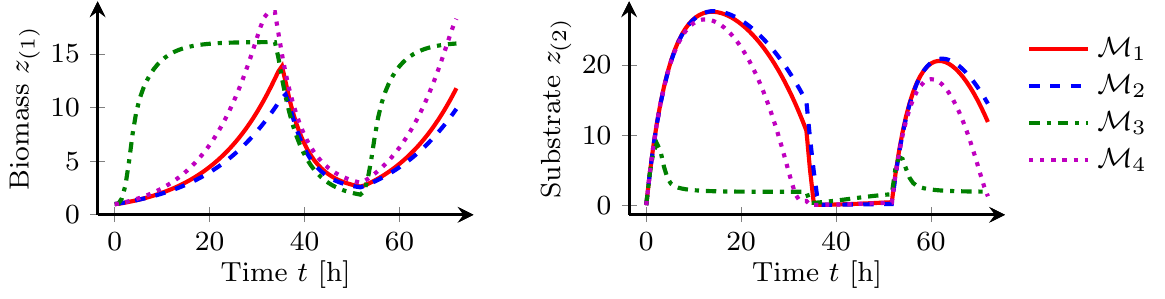}
	\vspace{-4mm}
	% \caption{Yeast fermentation case study model predictions for \cite{EspieMacchietto1989} optimal control inputs \new{$\mat U_\text{EM}^\ast$}.}
	\caption{\new{Predictions by the four models in \citeauthor{EspieMacchietto1989}'s (\citeyear{EspieMacchietto1989}) yeast fermentation case study, given optimal control inputs $\mat U_\text{EM}^\ast$.}}
	\label{fig:EspieMacchietto1989opt}
\end{figure}

\cite{EspieMacchietto1989} do not consider any uncertainty in their experimental design algorithm. Let $\mat U_\text{EM}^\ast$ denote the optimal control signal found by \cite{EspieMacchietto1989}. \figref{fig:EspieMacchietto1989opt} shows the corresponding model predictions. We use the model parameter values $\vec \theta_i$ reported in \cite{EspieMacchietto1989}. If we perturb the elements of $\mat U_\text{EM}^\ast$ by 1.5\%, 5\% and 10\% and use the perturbed inputs as the initial point for the optimisation algorithm, we retrieve $\mat U_\text{EM}^\ast$ again.

%
% Comparison to Chen & Asprey (2003) results
%
%\subsection{Comparison to \cite{ChenAsprey2003}}
Let us add parametric uncertainty and a design criterion accounting for measurement noise. \cite{ChenAsprey2003} discriminate between the two most similar models: $\M_1$ and $\M_2$. They find 20 optimal measurement time instances $\Tmeas$ and an optimal control signal $U_\text{CA}^\ast$ consisting of 5 piece-wise constant sections. Using the \cite{ChenAsprey2003} divergence measure, we find objective function value 2601 for their solution. For the same measurement time instances and control switch time instances, we find a solution $U_\text{new,1}^\ast$ with objective function value 2805, i.e.\ a 7.8\% larger divergence. \remove{W}\new{Since the problem formulation is identical, w}e assume this discrepancy is due to the improvement in optimisation solvers after 20 years.

\begin{figure}[!t]
    \centering
	\includegraphics[width=0.98\textwidth]{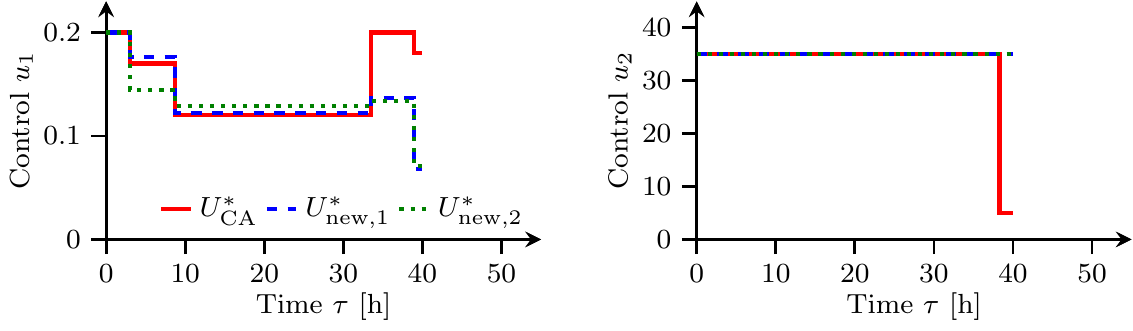}
	\vspace{-4mm}
    \caption{\new{Optimal control inputs} $U_\text{CA}^\ast$ from \cite{ChenAsprey2003}, our solution $U_\text{new,1}^\ast$, and our solution $U_\text{new,2}^\ast$ for the case of added process noise. \new{We see that adding process noise affects the optimal control signal.}}
    \label{fig:ChenAsprey2003Comparison}
\end{figure}

We add process noise with variance $\vec \Sigma_x \defeq 0.01 \cdot \mat I$\remove{, i.e.~the }\new{ (}same order of magnitude as \new{the }measurement noise\new{)}. %The optimal control signal $U_\text{CA}^\ast$ of \cite{ChenAsprey2003} yields a divergence of 244. 
\new{\citeauthor{ChenAsprey2003}'s (\citeyear{ChenAsprey2003}) optimal control signal $U_\text{CA}^\ast$ yields a divergence of 244 with the new (noisier) problem formulation.}
\figref{fig:ChenAsprey2003Comparison} shows our solution $U_\text{new,2}^\ast$ that yields a divergence of 263,\remove{ i.e.} an 8\% larger divergence. \figref{fig:ChenAsprey2003Comparison} shows that $U_\text{new,1}^\ast \neq U_\text{new,2}^\ast$, which means that adding process noise affects the optimal control signal.

%
% Mean and cone constraints
%
\subsection{Comparing Constraints}
\label{sec:compare_path_constraints}
We compare the results of solving the design of experiments optimisation problem in \eqref{eq:design_problem_formulation} using \remove{(i)}\new{either} a mean constraint, or \remove{(ii) }a cone chance constraint with $\alpha=2$ standard deviations margin. We compare the constraints' performance in terms of the number of constraint violations, how severe the violations are, and the number of experiments required for successful model discrimination. This requires choosing a model discrimination criterion. The model discrimination literature describes several different discrimination criteria
%, e.g.\ ranking based on normalised model likelihoods \citep{BoxHill1967}, the $\chi^2$ test \citep{BuzziFerraris1983}, and ranking based on Akaike weights \citep{Michalik2010}. 
\citep{BoxHill1967,BuzziFerraris1990,Michalik2010}.
Following \cite{BuzziFerraris1990}, we consider the $\chi^2$ test. 
The weighted squared residuals $\delta_{\timeind}^{(i)} 
= 
( \vec y_{\timeind} - \mat H_i \vec \mu_{\timeind}^{(i)} )\T
( \mat H_i \vec \Sigma_{\timeind}^{(i)} \mat H_i\T + \vec \Sigma_y )^{-1} 
( \vec y_{\timeind} - \mat H_i \vec \mu_{\timeind}^{(i)} )$ 
%\begin{align*}
%	\delta_{\timeind}^{(i)} &= 
% 	\left( \vec y_{\timeind} - \mat H_i \vec \mu_{\timeind}^{(i)} \right)\T
% 	\left( \mat H_i \vec \Sigma_{\timeind}^{(i)} \mat H_i\T + \vec \Sigma_y \right)^{-1} 
% 	\left( \vec y_{\timeind} - \mat H_i \vec \mu_{\timeind}^{(i)} \right) \,,
%\end{align*}
should be $\chi^2$-distributed. The $\chi^2$ score is 1 minus the $\chi^2$ cumulative distribution at $\sum_{\timeind \in \Tmeas} \delta_{\timeind}^{(i)}$ with $|\Tmeas| \times \dimy - \dimpi$ degrees of freedom. Models are inadequate if their $\chi^2$ score is below some threshold, \remove{e.g.}\new{such as}~\new{$1\textsc{e-}3$}. 
%The $\chi^2$ test is more conservative than ranking based on log-likelihood or Akaike weights, which may result in more experiments but also less risk of selecting an incorrect model.

Let $\bar{z}_2$ be a constraint upper bound on the substrate concentration, such that we wish to satisfy $z_{\timeind,(2)} \leq \bar{z}_2$ for $\timeind=\lbrace 1, \dots, T \rbrace$. For the simulations, the upper bound takes a value $\bar{z}_2 \in \lbrace 7, 10, 15 \rbrace$. Measurements $\vec y_{1:T}$ are generated in each simulated experiment and used for model discrimination, with models deemed inadequate if their $\chi^2$ score is below \new{$1\textsc{e-}3$}. \tabref{tab:path_const_exp} shows the performance in 25 simulations of the different state constraint types (mean constraint and cone constraint) in terms of the average number of experiments needed for successful model discrimination; All simulations correctly identify Model $\M_1$ as the true data-generating model. At least for the test instance by \cite{ChenAsprey2003}, the additional conservatism of the cone constraints does not lead to\remove{ (i)} more required experiments or\remove{ (ii)} better model prediction performance.

\begin{table}[!t]
	\centering
	\begin{tabular}{c l | c c | c c c c c c c} %*{5}{>{\raggedleft\arraybackslash}p{2.2em}}  }
		& & \multicolumn{2}{c|}{\textbf{Experiments}} & \multicolumn{7}{c}{\textbf{Avg. num. of models}} \\
		& & \multicolumn{2}{c|}{\textbf{required}} & \multicolumn{7}{c}{\textbf{remaining after \#{}n exp.}} \\
		\textbf{Bound} & \textbf{Constraint} & Mean & Std
		& \hspace{1mm} & \#{}1 & \hspace{0.1mm} & \#{}2 & \hspace{0.1mm} & \#{}3 & \Tstrut\Bstrut\\ 
		\hline
		\multirow{2}{*}{7} & Mean & 2.02 & 0.14 & & 2.00 & & 1.02 & & 1 & \Tstrut\Bstrut\\
		& Cone & 2.42 & 0.50 & & 2.54 & & 1.42 & & 1 & \Tstrut\Bstrut\\ 
		\hline
		\multirow{2}{*}{10} & Mean & 2.04 & 0.20 & & 2.00 & & 1.04 & & 1 & \Tstrut\Bstrut\\
		& Cone & 2.16 & 0.37 & & 2.16 & & 1.16 & & 1 & \Tstrut\Bstrut\\
		\hline
		\multirow{2}{*}{15} & Mean & 2.12 & 0.33 & & 2.02 & & 1.12 & & 1 & \Tstrut\Bstrut \\
		& Cone & 2.02 & 0.14 & & 2.02 & & 1.02 & & 1 & \Tstrut\Bstrut\\
		\hline \hline
		15 & Cone & 2.12 & 0.33 & & 2.08 & & 1.12 & & 1 & \Tstrut\Bstrut
	\end{tabular}
	\caption{Average number of experiments required (with standard deviation) to discard the incorrect models in 25 simulations of the yeast fermentation case study, with a mean or cone constraint (\secref{sec:constraints}) with an upper bound on $z_{\timeind,2}$. The right-most columns show the average number of models (out of four) that pass the $\chi^2$ test after 1, 2 or 3 experiments. The last row shows the result for the GP surrogate approach in \secref{sec:gp_results}. For the test instance by \cite{ChenAsprey2003}, the additional conservatism of the cone constraints does not lead to \remove{(i) }more required experiments or \remove{(ii) }better model prediction performance.}
	\label{tab:path_const_exp}
\end{table}
%	\todo{it sounds weird: Chen-test instance (in caption)}

But the following results indicate that the conservatism of the cone constraints do indeed increase the safety of the control signal by lowering the violation. We generate the experimental measurements $\vec y_{1:T}$ from one particular realisation of initial state, control inputs and measurement noise. If the constraints are not violated for this particular noise realisation, they may still 
%be likely to 
be violated for other experiments generated using the same optimal control signal $\hat{\vec u}_{1:T}^\ast$. Therefore, we use Monte Carlo sampling to better assess the safety of a control signal.

\new{We generate data by evaluating model $\M_1$, with ``true'' model parameter values $\vec \theta^{(0)}$ from \citet{ChenAsprey2003}.} Let $\M_0$ denote the data-generating model\remove{ ($\M_1$ with ``true'' model parameter values)}\new{, with transition function $f_0(\cdot,\cdot) = f_1(\cdot, \cdot, \vec \theta^{(0)})$}. Assume we are studying an optimised control sequence $\hat{\vec u}_{0:T-1}^\ast$. We generate a large number $N_\text{sim}=100$ of noisy control sequences $\vec u_{0:T-1,n}$, $n=1,\dots,N_\text{sim}$ by drawing samples $\vec u_{\timeind,n} \sim \N(\hat{\vec u}_{\timeind}^\ast, \vec \Sigma_u)$. For each control sequence, we sample a random initial state $\vec x_{0,n} \sim \N([1,\,0.01]\T, \vec \Sigma_0)$ and process noise sequence $\vec w_{0:T-1,n}$. The control sequences, initial states and process noise sequences are used with \remove{the true model }$\M_0$ to generate corresponding sequences of observed states $\vec z_{1:T,n}$. Let $\mat Z = \lbrace \vec z_{1:T,1}, \dots, \vec z_{1:T,N_\text{sim}} \rbrace$ denote the set of observed state sequences. Next, let $\mat Z_\text{viol} = \lbrace \vec z_{1:T} \,|\, \vec z_{1:T} \in \mat Z \,\wedge\, \exists \timeind: z_{\timeind,(2)} > \bar{z}_2 \rbrace$ be the set of observation sequences for which the constraint $z_{\timeind,(2)} \leq \bar{z}_2$ is violated for at least one time step $\timeind$. The experiment \textit{violation level} is the relative size $|\mat Z_\text{viol}| / N_\text{sim}$. A violation level of 0\,\% means that a given control signal is apparently safe, while a violation level of 100\,\% means that a control signal is almost guaranteed to result in constraint violations.

\begin{figure}
	\centering
	\includegraphics[width=0.98\textwidth]{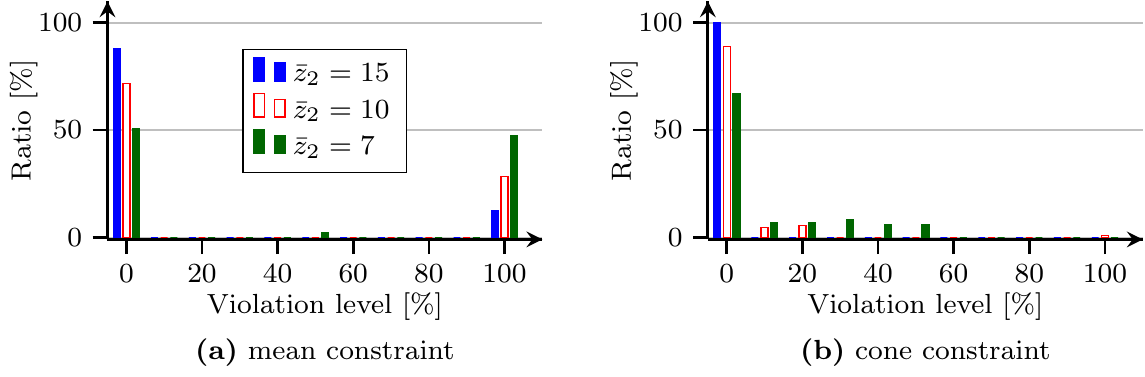}
	\vspace{-4mm}
	\caption{The violation level is the ratio of Monte Carlo simulations using optimised control signals that result in constraint violations $z_{\timeind,(2)} > \bar{z}_2$. The vertical axes show the ratio of optimised control signals for each violation level (horizontal axes) for a given constraint.}
	\label{fig:ctest_column_chart}
\end{figure}

%The histograms in 
%The plots in
\figref{fig:ctest_column_chart} shows the ratio of experiments, i.e.\ optimised control signals, at each violation level.
Note:\remove{ (i)} The cone constraint results in \textit{fewer} constraint violations than the mean constraint;\remove{ (ii)} The cone constraint results in \textit{less severe} constraint violations than the mean constraint;\remove{ (iii)} The number of constraint violations increases as the upper bound shrinks. These observations are in line with expectations.

%
% Control uncertainty
%
\subsection{Effect of Control Input Uncertainty}
\label{sec:control_uncertainty}
Problem \eqref{eq:design_problem_formulation} assumes several covariance matrices are known. In many applications, the exact covariances are unknown and have to be approximated. We study the effect of under- or overestimating the covariance size by varying the control input covariance $\vec \Sigma_u$. More specifically, let the models $\M_i$ assume the control input $\vec u_{\timeind} \sim \N(\hat{\vec u}_{\timeind}, \hat{\vec \Sigma}_u)$. We generate experimental data \new{by evaluating $\M_0$ (same as in \secref{sec:compare_path_constraints}), }using control inputs $\vec u_{\timeind} \sim \N(\hat{\vec u}_{\timeind}, \vec \Sigma_u)$. \remove{In the simulations, t}\new{T}he \emph{modelled} covariance $\hat{\vec \Sigma}_u$ and the \emph{true} covariance $\vec \Sigma_u$ \remove{are assigned}\new{take} values $\vec \Sigma_\text{small} = \diag(1\textsc{e-}8,\,1\textsc{e-}4)$ \remove{and}\new{or} $\vec \Sigma_\text{large} = \diag(1\textsc{e-}4,\,1\textsc{e-}2)$.
%$\diag(1\textsc{e-}8,\,1\textsc{e-}4)$ (denoted ``small'') and $\diag(1\textsc{e-}4,\,1\textsc{e-}2)$ (denoted ``large'').
%denoted ``small'' and ``large'', with
%\begin{subequations}
%\label{eq:input_covariances}
%\begin{align}
%    \label{eq:input_covariances}
%	\textbf{small:}\quad %&
%	\diag(1\textsc{e-}8,\,1\textsc{e-}4) \,, 
%	\quad \quad %\\
%	\textbf{large:}\quad %&
%	\diag(1\textsc{e-}4,\,1\textsc{e-}2) \,.
%\end{align}
%\end{subequations}
There are four combinations of small and large modelled and true control covariances. The resulting scenarios can be described as (i) correctly modelled uncertainty $\hat{\vec \Sigma}_u) = \vec \Sigma_u$, (ii) underestimated uncertainty $| \hat{\vec \Sigma}_u)| \leq |\vec \Sigma_u|$, and (iii) overestimated uncertainty $| \hat{\vec \Sigma}_u| \geq |\vec \Sigma_u|$.

\tabref{tab:path_u_exp} shows the result of 25 simulations of the yeast fermentation case study with the different modelled and true control covariances.
\begin{table}[!t]
	\centering
	\begin{tabular}{c c | c c | c c c | c}
		\multicolumn{2}{c|}{\textbf{Control}} & \multicolumn{2}{c|}{\textbf{Experiments}} & \multicolumn{3}{c|}{\textbf{Model}} & \textbf{Cone} \\
		\multicolumn{2}{c|}{\textbf{covariance}} & \multicolumn{2}{c|}{\textbf{required}} & \multicolumn{3}{c|}{\textbf{discrimination}} & \textbf{constraint} \\
		$\hat{\vec \Sigma}_u$ & $\vec \Sigma_u$ & Mean & Std 
		& Succ. & Fail. & Inconcl. & \textbf{violations} \Tstrut\Bstrut\\ 
		\hline
		small & small & 2.04 & 0.20 & 100\,\% & 0\,\% & 0\,\% & 0\,\% \Tstrut\Bstrut\\
		small & large & 2.05 & 0.22 & 84\,\% & 0\,\% & 16\,\% & 4\,\% \Tstrut\Bstrut\\
		large & small & 2.58 & 0.97 & 96\,\% & 0\,\% & 4\,\% & 0\,\% \Tstrut\Bstrut\\
		large & large & 2.21 & 0.66 & 96\,\% & 0\,\% & 4\,\% & 0\,\% \Tstrut\Bstrut\\
	\end{tabular}
	\vspace{-1mm}
	\caption{The first set of columns shows the modelled control covariance $\hat{\vec \Sigma}_u$ and the true control covariance $\vec \Sigma_u$ used for generating experimental data, with $\vec \Sigma_\text{small} = \diag(1\textsc{e-}8,\,1\textsc{e-}4)$ and $\vec \Sigma_\text{large} = \diag(1\textsc{e-}4,\,1\textsc{e-}2)$. 
	The second set of columns shows the average number of experiments required for successful model discrimination in 25 yeast fermentation case study simulations. The third set of columns shows the ratio of successful, failed or inconclusive model discrimination. The last column shows the ratio of simulations with cone constraint violations.}
	\label{tab:path_u_exp}
\end{table}
A cone constraint is enforced with an upper bound $\bar{z}_2=15$
%\,g/L on the substrate concentration 
(see \secref{sec:compare_path_constraints}). As expected, a correctly modelled small control covariance yields the best result in terms of average number of experiments required and the model discrimination success rate. A large modelled control covariance increases the average number of required experiments, and decreases the success rate. Model discrimination \emph{fails} if an incorrect model is identified as the data-generating model, and is \emph{inconclusive} if the experimental budget (maximum number of allowed experiments) is exhausted or the $\chi^2$-test discards all models as inadequate. These simulations never exhausted the experimental budget, i.e.\ all inconclusive model discrimination instances are due to all models being deemed inadequate. The rate of inconclusive model discrimination is significantly higher when the true control covariance is underestimated, and we have a cone constraint violation. Hence, we are punished less for conservative estimates of the control covariance than for overly optimistic estimates.

%
% GP experiments
%
\subsection{Black-Box Transition Functions}
\label{sec:gp_results}
\new{The next level of difficulty is black-box transition functions. To solve the design of experiments problem with black-box transition functions we introduce our GP surrogate models.}\remove{Next we computationally compare the GP surrogate approach to the \secref{sec:compare_path_constraints} analytic results. This approach uses GP surrogate predictions during design of experiments.} \new{Here we encounter an issue with the model parameter prior covariances.} Each simulation starts with no initial data and\remove{ a} relatively uninformed model parameter \remove{distribution}\new{priors}. \remove{This model parameter distribution designs the first experiment. After executing an experiment, we update the model parameter distribution and then compute the models' $\chi^2$ score during model discrimination. For the analytical transition functions i}\new{I}n the \secref{sec:compare_path_constraints} state constraint test and the \secref{sec:control_uncertainty} control uncertainty test,\new{ with analytical transition functions,} we use the \appref{app:yeast_fermentation_cs} model parameter prior\new{ covariance $\vec \Sigma_{\theta,i,0} = 0.05 \cdot \mat I$}. However, for\new{ the} GP surrogates, which\new{ additionally} incorporate model prediction uncertainty, th\remove{e model parameter prior}\new{is covariance} is too large and the \remove{uncertainty}\new{models' predictive distributions} cannot satisfy the constraint $\bar{z}_2=15$\new{ with sufficient probability}. We solve this by reducing the\remove{ variance in the} model parameter prior\new{ covariance} to $\vec \Sigma_{\theta,i,0} = 1\textsc{e-}4 \cdot \mat I$ for the GP surrogates. On the one hand, this means the\new{ parameter} prior\remove{ parameter estimate} may be overly confident \remove{and there is a higher probability}\new{with increased risk} of constraint violations in the GP surrogate tests\remove{, b}\new{. B}ut\remove{ also} the optimisation is more likely to converge on a feasible first experiment in each simulation.

\new{We compare the GP surrogate approach to the \secref{sec:compare_path_constraints} analytic results.} The last row of \tabref{tab:path_const_exp} shows the GP\new{'s}\remove{ surrogate approach} performance in terms of average number of experiments required for successful model discrimination in 25 simulations of the yeast fermentation case study. Model $\M_1$ was correctly identified as the true data-generating model in all simulations. The GP surrogate approach has a marginally worse performance than the analytical method.\new{ This is expected since the GP surrogate approach emulates the analytical method, but with more uncertain predictions due to the added covariance term in \eqref{eq:transition_dtss}.} \figref{fig:c_vs_gp_test} shows the average $\chi^2$ score evolution (with one standard deviation) for the four rival models in the yeast fermentation case study:
\begin{figure}[!t]
	\centering
	\includegraphics[width=0.98\textwidth]{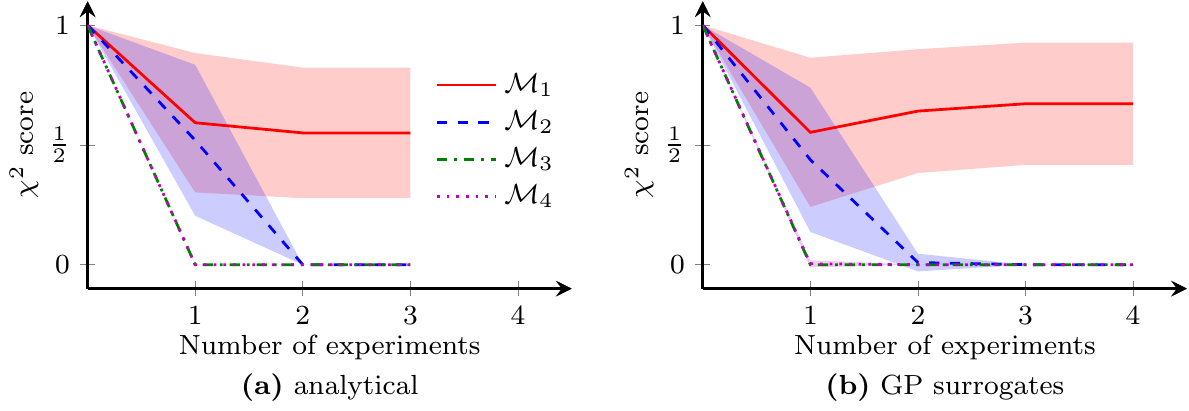}
	\vspace{-4mm}
	\caption{Evolution of the average $\chi^2$ scores (with one standard deviation) for the four rival models in the yeast fermentation case study, with experiments design using (a) the analytical approach, and (b) the GP surrogates approach.}
	\label{fig:c_vs_gp_test}
\end{figure}
\figref{fig:c_vs_gp_test}a\remove{ shows the average $\chi^2$ score} for the analytical approach with a cone constraint, and \figref{fig:c_vs_gp_test}b\remove{ the average $\chi^2$ score} using the GP surrogates approach. This illustrates the GP surrogates' marginally worse performance compared to the analytical method. Black-box models yield larger marginal model prediction uncertainty that results in higher $\chi^2$ scores\new{, which means, compared to non-black-box models, more experiments may be required for model discrimination}.

%
% Discussion
%
\section{Discussion}

Our work assumes that the black-box model component is the state transition function $f$. Our problem formulation writes all models as first-order models, i.e.\ for discrete-time models the state $\vec x_{\timeind+1}$ depends only on the state $\vec x_{\timeind}$, and for continuous-time models the first-order ordinary differential equation system have only the zeroth-order states $x(\time)$ on the right-hand side. All $n$th order models can be expressed as first-order models by introducing additional states.
We also assume the observed states $\vec z_{\timeind}$ are a linear combination of the states $\vec x_{\timeind}$. Since the transition function $f$ can be any non-linear function, using $\vec z_{\timeind} = \mat H \vec x_{\timeind}$ is only a minor restriction. 
%to the model types we consider for domains such as pharmaceuticals and chemical manufacturing.
%
State space models are also used 
%, where the states may be coordinates and velocities and the observed states are images \citep{Wahlstrom2015}, in which case
in settings where the mapping from state to observed state is highly non-linear, \remove{e.g.}\new{such as}~in pixels-to-torque problems \citep{Wahlstrom2015}. 
%In the more general case of $\vec z_{\timeind} = g(\vec x_{\timeind})$, with $g$ a non-linear function, 
Approximate moment matching can infer the predictive distribution for $\vec z_{\timeind}$ for this more general case $\vec z_{\timeind} = g(\vec x_{\timeind})$.

\new{A limitation of the analytical approach is the use of linear and Gaussian approximations for uncertainty propagation. This is done for analytical (and hence computational) tractability. It is important to be aware that linear approximation and GP surrogates introduce an error at each modelled time step, which may accumulate and grow significantly over longer time horizons. This may lead to poor model predictions, lower-than-expected experimental data informativeness, and constraint violations.}

\new{We assume the noise covariances $\vec \Sigma_x$ and $\vec \Sigma_y$ are known. In practice, estimating these covariances requires domain expertise. Our results in \secref{sec:control_uncertainty} show that conservative noise covariance estimates produce better results: The model noise covariances should provide reasonable upper bounds to the true noise variance.}

% \tabref{tab:dynamic_mapping_dimensionality} lists the input and output dimensionality of the GP mapping for both th\remove{e}\new{is paper's} approach \remove{in this paper }and \new{for }\cite{Olofsson2018_ICML}. Th\remove{is}\new{e latter} approach would not work for design of \emph{dynamic} experiments since \remove{(i) }the input dimensionality would be too high for accurate GP inference and \remove{(ii) }the output dimensionality\new{ (the number of GP surrogate models)} would \remove{require many GP surrogate models}\new{depend on $T$}. We reduce the input and output dimensionality by putting the GP prior on the state transition.
\removecmd{\begin{table}[!t]
	\centering
	\colorbox{red}{\begin{tabular}{c | c | c | c}
		%& \textbf{Black-Box} & \textbf{Input} & \textbf{Output} \\
		%\textbf{Approach} & \textbf{Mapping} & \textbf{dimensionality} & \textbf{dimensionality} \Tstrut\Bstrut\\ \hline
		\textbf{Approach} & \textbf{Mapping} & \textbf{Input dim.} & \textbf{Output dim.} \Tstrut\Bstrut\\ \hline
		\cite{Olofsson2018_ICML} & $\vec x_0, \vec u_{0:T-1}, \vec \theta\,\mapsto\,\vec z_{1:T}$ & $\dimx + \dimu \times T + \dimp$ & $\dimz \times T$ \Tstrut\Bstrut\\
		This work & $\vec x_{\timeind}, \vec u_{\timeind}, \vec \theta\,\mapsto\,\vec x_{\timeind+1}$ & $\dimx + \dimu + \dimp$ & $\dimx$ 
	\end{tabular}}
	\vspace{-2mm}
	\caption{\remove{Comparison of the input and output dimensionality different GP surrogate models have for different mappings.}}
	\label{tab:dynamic_mapping_dimensionality}
\end{table}}
% Table 5 lists the input and output dimensionality of the GP mapping for both the approach in this paper and \cite{Olofsson2018_ICML}. This approach would not work for design of dynamic experiments since (i) the input dimensionality would be too high for accurate GP inference and (ii) the output dimensionality would require many GP surrogate models. We reduce the input and output dimensionality by putting the GP prior on the state transition.
\new{\cite{Olofsson2018_ICML} use GP surrogate models for the mapping from inputs to observed output. With this paper's notation, this corresponds to the mapping $\vec x_0, \vec u_{0:T-1}, \vec \theta\,\mapsto\,\vec z_{1:T}$, with input dimensionality $\dimx + \dimu \times T + \dimp$ and output dimensionality $\dimz \times T$.}\remove{\tabref{tab:dynamic_mapping_dimensionality} lists the input and output dimensionality of the GP mapping for both the approach in this paper and \cite{Olofsson2018_ICML}.} This approach would not work for design of \emph{dynamic} experiments since \remove{(i) }the input dimensionality would be too high for accurate GP inference and \remove{(ii) }the output dimensionality\new{ (the number of GP surrogate models)} would \remove{require many GP surrogate models}\new{depend on $T$}. We \remove{reduce the input and output dimensionality by }put\remove{ting} the GP prior on the state transition\new{, corresponding to the mapping $\vec x_{\timeind}, \vec u_{\timeind}, \vec \theta\,\mapsto\,\vec x_{\timeind+1}$. This gives input dimensionality $\dimx + \dimu + \dimp$ and output dimensionality $\dimx$}. 

We consider an open-loop\new{ (offline)} control approach, where a designed experiment runs to completion before data is analysed. \cite{Galvanin2009} and \cite{DeLuca2016} consider a closed-loop\new{ (online)} approach for parameter estimation.
%, where the optimal control input sequence is updated online at time step $\timeind$ using data collected up until time step $\timeind-1$.
\remove{But in our setting, a}\new{A} closed-loop approach repeatedly solves the problem in \eqref{eq:design_problem_formulation}. Although theoretically possible for designing dynamic experiments for model discrimination, the computational cost\new{ of our approach} would likely be too high in practice.

%
% Conclusions
%
\section{Conclusions}

\new{Model discrimination finds mechanistic models that adequately describe and predict a system's behaviour. Mechanistic models are often needed in industry, for example to satisfy regulatory requirements.} We\new{ considered a wide range of problem uncertainty,} extended traditional analytic approaches for design of dynamic experiments using the \cite{Olofsson2018_ICML} methodology of replacing black-box models with GP surrogates\remove{. We also considered a wider range of problem uncertainty}, and unified the literature contributions for discrete- and continuous-time models. Literature comparisons show that our \remove{models}\new{method} reduce\new{s} to previously acquired results. Both this paper and \cite{Olofsson2018_ICML} present the GP surrogate approach as hybridising analytical and data-driven \new{experimental design }approaches\remove{ to optimal design of experiments}. Condensing the range of possibilities, the GP surrogate approach is closer to the analytical than the fully data-driven approach. But we imagine a spectrum of different trade-offs between accuracy and computational complexity. The posteriors of more advanced GP models may not have closed form expressions \citep{Salimbeni2018}, so we may require sampling to approximate \new{the }predictive distributions. Hybrid design of experiments approaches using such advanced GP surrogates therefore lie closer to the \remove{original }data-driven approaches.\remove{ Model discrimination finds mechanistic models that adequately describe and predict a system's behaviour. Mechanistic models are often needed in industry, e.g.~to satisfy regulatory requirements. In future, more models may be completely data-driven.} An alternative, hybrid approach could combine mechanistic modelling and data-driven learning for data-driven models with physically meaningful embeddings \citep{Saemundsson2019}. The model discrimination challenge becomes discerning the effect of the mechanistic versus the data-driven model part.

\begin{small}
\vspace{3mm}
\noindent
\new{\textbf{Author contributions} 
Conceptualisation, S.O. and R.M.; %
Methodology / Software, S.O. and E.S.S.; %s
% Methodology, S.O. and E.S.S.; %
Validation / Formal Analysis / Investigation / Data Curation, S.O.; %
% Methodology, S.O.; %
% Validation, S.O.; %
% Formal Analysis, S.O.; %
% Investigation, S.O.; %
% Data Curation, S.O.; %
Writing (Original Draft Preparation) / Vizualisation, S.O.; % 
% Writing–Original Draft Preparation, S.O.; % 
Writing (Review and Editing), all; %
% Visualization, S.O.; %
Supervision, Ad.M., Al.M, M.P.D. and R.M.; %
Project Administration / Funding Acquisition / Resources, Al.M. and R.M. %
% Resources, A. Mi. and R.M.; %
% Project Administration, A. Mi. and R.M.; %
% Funding Acquisition, A. Mi. and R.M.
}

\vspace{2mm}
\noindent
\new{\textbf{Funding} This work has received funding from the European Union's Horizon 2020 research and innovation programme under the Marie Skl{}odowska-Curie grant agreement no. 675251, an EPSRC Research Fellowship (EP/P016871/1), and the Imperial Data Science Institute.}

\vspace{2mm}
\noindent
\new{\textbf{Code availability} We generated all results using our open-source Python package for design of experiments, \textit{doepy} (\url{https://github.com/scwolof/doepy}).}

\vspace{2mm}
\noindent
\textbf{Conflict of interest} The authors declare that they have no conflict of interest.

\vspace{2mm}
\noindent
\new{\textbf{Ethics approval / Consent to participate / Consent for publication} Not applicable.}
\vspace{-6mm}
\bibliography{ref.bib}
\end{small}

% Appendices
\pagebreak
\appendix

\section{The \cite{Bania2019} Case Study}
\label{app:bania_models}

\new{The \cite{Bania2019} case study considers three rival, linear continuous-time models $\M_i$, $i \in \lbrace 1, 2, 3 \rbrace$, of the form}
\begin{align*}
    \alignedbox{
	\M_i:\quad \left\lbrace
	\begin{aligned}
		\left. \frac{\mathrm{d} x}{\mathrm{d}\time} \right|_{\time}
		&= \mat A_i x(\time) + \mat B_i u(\time) + \mat C_i w (\time) \,,\\
		y_\timeind  &= [1,\,0,\,\dots,0] \, x(\time_\timeind) + 0.05 \cdot v_\timeind \,,
	\end{aligned}
	\right.
	}
\end{align*}
\new{where $x(\time)$ denotes the state at time $\time$, $u(\time)$ denotes the control input and $w$ is the process noise given by a Wiener process with variance 1. Measurements $y_k$ are taken at discrete time points $\time_\timeind$, $\timeind=1, \dots, T$ with Gaussian-distributed measurement noise $v_\timeind \sim \N(0, 1)$.} The matrices $\mat A_i$, $\mat B_i$ and $\mat C_i$ \remove{in the \cite{Bania2019} case study in \eqref{eq:dode_example1} }are defined as
\begin{align*}
	\M_1 &: \quad \mat A_1 = -1 \,, \quad \mat B_1 = 1 \,, \quad \mat C_1 = 0.05 \,,\\
	\M_2 &: \quad
	\mat A_2 = \begin{bmatrix}
		\phantom{-}0 & \phantom{-}1\,\, \\ 
		-3 & -2.5\,\,
	\end{bmatrix} \,, \quad
	\mat B_2 = \begin{bmatrix}
		\,0\,\, \\ \,3\,\,
	\end{bmatrix} \,, \quad
	\mat C_2 = \begin{bmatrix}
		\,0\,\, \\ \,0.05\,\,
	\end{bmatrix} \,, \\
	\M_3 &: \quad
	\mat A_3 = \begin{bmatrix}
		\phantom{-}0 & \phantom{-}1 & \phantom{-}0\,\, \\ 
		-3 & -3.5 & \phantom{-}1\,\, \\ 
		\phantom{-}0 & \phantom{-}0 & -10\,\,
	\end{bmatrix} \,, \quad
	\mat B_3 = \begin{bmatrix}
		\,0\, \\ \,0\, \\ \,30\,\,
	\end{bmatrix} \,, \quad
	\mat C_3 = \begin{bmatrix}
		\,0\, \\ \,0\, \\ \,0.05 \,\,
	\end{bmatrix} \,.
\end{align*}
The initial latent state is $\vec x_0^{(i)} = [0,\,\dots,\,0]\T$\remove{, and the process noise $w(\time) \sim \N(0, 1)$ and the measurement noise $v_\timeind \sim \N(0, 1)$ are Gaussian distributed}.

\section{Analytical Design of Dynamic Experiments}
\label{app:existing_work}

Extensions of the analytical methods exist for design of dynamic experiments. 
\cite{EspieMacchietto1989} %and \cite{AspreyMacchietto2000} 
consider discrimination between multiple analytic continuous-time models and formulate an optimal control problem. \cite{EspieMacchietto1989} compare the results of using optimal constant control inputs \textit{versus} an optimal dynamic control input. %\cite{AspreyMacchietto2000} discuss accounting for model parameter uncertainty by taking the expected value over the design criterion, or by considering worst-case parameter realisations. 
\cite{ChenAsprey2003} also consider continuous-time models and use a Laplace approximation for the model parameter covariance and linear propagation of the Gaussian model parameter uncertainty to approximate the marginal predictive distributions.

\cite{SkandaLebiedz2010} assume Gaussian measurement noise to derive an expression for the Kullback-Leibler (KL) divergence between the predictive distributions of two rival models (with the same number of states). They include the measurement time points $\Tmeas$ as variables in the optimisation problem, together with the initial state $x(0) = \vec x_0$ and control inputs $\vec u_{0:T-1}$. The control inputs \cite{SkandaLebiedz2010} consider are additive perturbations to the state, and they assume the states cannot be measured and perturbed in the same time step. \cite{SkandaLebiedz2013} extend the setup of \cite{SkandaLebiedz2010} by considering model parameter uncertainty. They propose a robust optimisation formulation
\begin{align*} 
    \argmax_{\substack{\Tmeas \\ \vec x_0 \in \X \\ \vec u_{0:T-1} \in \U }} 
    \min_{\substack{i,j \in \lbrace 1, \dots, M \rbrace \\ i \neq j}} 
    \min_{\substack{ \vec \theta_i \in \vec \Theta_i \\ \vec \theta_j \in \vec \Theta_j }}  
    %U(\vec x_0, \vec u_{1:T}, \vec \theta_i, \vec \theta_j) \,,
    \sum_{\timeind \in \mathcal{T}_\mathrm{meas}} \mathrm{KL}\left[ p(\vec y_{\timeind} \,|\, \vec \theta_i ) \,\|\, p( \vec y_{\timeind} \,|\, \vec \theta_j) \right]
\end{align*}
subject to constraints, with model parameter spaces $\vec \Theta_i$ and $\vec \Theta_j$ and $p(\vec y_{\timeind}\,|\,\vec \theta_i)$ denoting the predictive distribution at time step $\timeind$ given model $i$ with parameter values $\vec \theta_i$.

None of \cite{EspieMacchietto1989}, %\cite{AspreyMacchietto2000}, 
\cite{ChenAsprey2003}, or \cite{SkandaLebiedz2010,SkandaLebiedz2013} consider process noise or uncertainty in the initial state $x(0) = \vec x_0$ or control signal $u(\time)$. Nor do any of them, when solving the optimisation problem subject to dependent variable constraints on the observed states $\vec z_{1:T}$, account for the uncertainty in the observed states $\vec z_{1:T}$ predictions.

\citet{CheongManchester2014a} consider non-parametric linear discrete-time systems with process noise (but no separate measurement noise) and uncertainty in the initial states $\vec x_0$. For optimising the control signal they consider design criteria based on the pairwise difference in models' score in the $\chi^2$ goodness-of-fit test.
%\citet{CheongManchester2014b} extend this approach by deriving a model discrimination control law for model predictive control. 
Though 
%\citet{CheongManchester2014a,CheongManchester2014b} 
\cite{CheongManchester2014a}
consider dependent variable constraints in the observed states $\vec z_{1:T}$, they do not account for the uncertainty in the observed states $\vec z_{1:T}$ predictions.

\cite{Streif2014} and \cite{Mesbah2014} look at cases of two rival non-linear models $\M_1$ and $\M_2$ with multiplicative measurement noise
\begin{align*}
	\M_i:\quad \left\lbrace
	\begin{aligned}
		\left. \frac{\mathrm{d} x(\time)}{\mathrm{d}\time} \right|_{\time} 
		&= f_i(x(\time), u(\time), \vec \theta_i) \,,\\
		z(\time) &= g_i(x(\time), u(\time), \vec \theta_i) \,,\\
		\vec y_{\timeind} &= \diag(\vec 1 + \vec w_{\timeind}) \vec z_{\timeind}
	\end{aligned}
	\right.
\end{align*}
where $f_i$ and $g_i$, $i \in \lbrace 1,2 \rbrace$, are polynomial functions. They consider uncertainty in the initial states and model parameters using polynomial chaos expansions, from which they compute higher moments of the predictive distributions---``a computationally formidable task'' according to \cite{Streif2014}. They discretise the control signal and solve the design problem by minimising the norm of the control signal such that the divergence between the predictive distributions is greater than or equal to some threshold value. The divergence can be computed using the predictive distributions' higher moments \citep{Streif2014} or through Markov Chain Monte Carlo integration \citep{Mesbah2014}.

\cite{KeesmanWalter2014} look at continuous-time models of the kind $\tfrac{\mathrm{d}}{\mathrm{d}\time} y(\time) = f(y(\time)) + b u(\time)$. They define the Hamiltonian and from this derive an optimal control law in closed form for two rival models. This requires gradient information of \emph{at least} the first order. They do not account for parametric uncertainty, process noise or measurement noise.

\cite{Bania2019} consider non-parametric linear discrete-time models with process and measurement noise. By looking at the mutual information between choice of model and observed output, they derive an optimisation formulation based on minimising the probability of selecting the wrong model. They mention how to extend their approach to non-linear models.

%\section{State Transition}
%\label{app:latent_state_prediction}

%
% Continuous-Time Models
%
\section{Continuous-Time State Space Models}
\label{app:cont_model}
The continuous-time state space models is described by
\begin{align*}
    %\label{eq:ctss}
    \M:\quad
    \left\lbrace
    \begin{aligned}
        \left. \dfrac{\mathrm{d} x}{\mathrm{d} \time} \right|_{\time}
        %\frac{\mathrm{d}}{\mathrm{d}\time} x(\time) 
        &= f_i ( x(\time), u(\time), \vec \theta ) + w(\time) \,, \\
        x(\time_0) &\defeq \vec x_0 \,,
        \\
        z(\time) &= \mat H \, x (\time) \,, 
        \\
        \vec y_{\timeind} &= z(\time_{\timeind}) + \vec v_{\timeind} \,,
    \end{aligned}
    \right.
\end{align*}
with state $x(\time)$, control input $u(\time)$, process noise distribution $w(\time) \sim \N(\vec 0, \vec \Sigma_x)$. The control input $u(\time_{\timeind})$ may be a continuous function of the time $\time$, but we will assume it is piece-wise constant. Model $\M$'s state prediction at time step $\timeind$ is given by %the solution $\vec x_{\timeind} = x(\time_{\timeind})$ at time $\time_{\timeind}$.
\begin{align*}
    \vec x_{\timeind}
    =
    \vec x_{\timeind-1}
    + 
    \int_{\time_{\timeind-1}}^{\time_{\timeind}} f (x(\time), \,\vec u_{\timeind-1}, \,\vec \theta ) \mathrm{d}\time + \vec w_{\timeind}
\end{align*}
where $w(\time) \sim \N(\vec 0, (\time_{\timeind} - \time_{\timeind-1}) \vec \Sigma_x)$.

Let $\tilde{x}(\time)$ denote the continuous concatenated state, control input and model parameters $\tilde{x}(\time) = [x(\time)\T, u(\time)\T, \vec \theta\T]\T$, with Gaussian distribution $\tilde{x}(\time) \sim \N(\tilde{\mu}(\time),\tilde{\Sigma}(\time))$, and let $\tilde{\mu}_f$ denote the concatenated transition function
\begin{align*}
	\tilde{\mu}_f (\time) &= \left[ \mu_f(\tilde{\mu}(\time))\T ,\, \vec 0 ,\, \vec 0 \right]\T \,, \quad \tilde{\mu}_f (\time) \in \R^{\dimx + \dimu + \dimp} \,.
\end{align*}
We find the state prediction $\vec x_{\timeind} \sim \N(\vec \mu_{\timeind}, \vec \Sigma_{\timeind})$ at time step $\timeind$ by extracting the corresponding elements from $\tilde{\mu}(\time_{\timeind})$ and $\tilde{\Sigma}(\time)$ (see \eqref{eq:variable_dependence}) which we compute by solving the following system of ordinary differential equations
\begin{align*}
	%\label{eq:transition_ctss}
	\begin{split}
    \left\lbrace
    \begin{aligned}
        % dM
        \left. \dfrac{\mathrm{d} \tilde{\mu}}{\mathrm{d} \time} \right|_{\time}
        %\frac{\mathrm{d}}{\mathrm{d} \time}\tilde{\mu}(\time) 
        &= \tilde{\mu}_f(\time) \,, \\
        % dS
        \left. \dfrac{\mathrm{d} \tilde{\Sigma}}{\mathrm{d} \time} \right|_{\time}
        %\frac{\mathrm{d}}{\mathrm{d} \time} \tilde{\Sigma}(\time) 
        &=
        \nabla_{\tilde{\mu}(\time)} \tilde{\mu}_f \tilde{\Sigma}(\time) + \tilde{\Sigma}(\time) \left( \nabla_{\tilde{\mu}(\time)} \tilde{\mu}_f \right)\T
        + \diag(\Sigma_f (\tilde{\mu}(\time)) + \vec \Sigma_x,\, \vec 0,\, \vec 0) \,, \\
        % Initial values
        \tilde{\mu}(\time_{\timeind-1}) &\defeq \tilde{\vec \mu}_{\timeind-1} \,,\\
        \tilde{\Sigma}(\time_{\timeind-1}) &\defeq \tilde{\vec \Sigma}_{\timeind-1} \,. \\
    \end{aligned}
    \right.
	\end{split}
\end{align*}
Derivatives of $\vec \mu_{\timeind+1}$ and $\vec \Sigma_{\timeind+1}$ with respect to $\vec \mu_{\timeind}$, $\vec \Sigma_{\timeind}$ and $\hat{\vec u}_{\timeind}$ are calculated by integrating over the chain rule, and require second-order derivative information of $f$ or the GP prediction.

For continuous-time models we may wish to optimise the measurement time instances $\Tmeas$. In this case a minimum amount of time between measurements need to be enforced during optimisation. Let the absolute difference in time between two measurement time points $\time_{\timeind}$ and $\time_{\timeind'}$ be lower-bounded by $\Delta_{\time} \geq 0$
\begin{align}
	\label{eq:time_abs_diff_constraint}
	\left| \time_{\timeind} - \time_{\timeind'} \right| \geq \Delta_{\time} \,,\quad \forall \time_{\timeind},\time_{\timeind'} \in \Tmeas \,.
\end{align}
This constraint is non-convex. To simplify the problem formulation, we introduce additional constraints to maintain a fixed order of the measurement time points, and reformulate \eqref{eq:time_abs_diff_constraint} in convex form as
\begin{align*}
	\forall \time_{\timeind},\time_{\timeind'} \in \Tmeas \,:\quad
	\begin{cases}
		\time_{\timeind} - \time_{\timeind'} - \Delta_{\time} \geq 0 \,, & \timeind \geq \timeind' \,, \\
		\time_{\timeind'} - \time_{\timeind} - \Delta_{\time} \geq 0 \,, & \timeind < \timeind' \,.
	\end{cases}
\end{align*}
using the constraints format in \eqref{eq:design_problem_formulation}.

%
% Discrete-Time with Delta-Transition
%
\section{Discrete-Time Model with $\Delta$-Transition}
\label{app:disc_delta_model}
The discrete-time state space model with a $\Delta$-transition is described by
\begin{align*}
    \M:\quad
    \left\lbrace
    \begin{aligned}
        \vec x_{\timeind+1}
        &= \vec x_{\timeind} + f ( \vec x_{\timeind}, \vec u_{\timeind}, \vec \theta ) + \vec w_{\timeind} \,, \\
        \vec z_{\timeind} &= \mat H \vec x_{\timeind} \,, \\
        \vec y_{\timeind} &= \vec z_{\timeind} + \vec v_{\timeind} \,,
    \end{aligned}
    \right.
\end{align*}
with process noise $\vec w_{\timeind} \sim \N(\vec 0, \vec \Sigma_x)$. The discrete-time model with a $\Delta$-transition follows from an Euler discretisation of continuous-time dynamics \citep[Ch.~2]{eulermethodbook}. Using a first-order Taylor expansion of $\mu_f(\tilde{\vec x})$ around $\tilde{\vec x}_{\timeind-1} = \tilde{\vec \mu}_{\timeind-1}$, the mean and variance of the state at time step $\timeind \geq 1$ in \eqref{eq:transit_state_mean} are approximately given by
\begin{align*}
	%\label{eq:transition_dtssDelta}
	\begin{split}
    \vec \mu_{\timeind} &\approx \vec \mu_{\timeind-1} + \mu_f(\tilde{\vec \mu}_{\timeind-1})
    \,, \\
	\vec \Sigma_{\timeind} 
	&\approx
	\nabla_{\tilde{\vec x}_{\timeind-1}} \vec \mu_{\timeind} 
	\tilde{\vec \Sigma}_{\timeind-1} 
	\left( \nabla_{\tilde{\vec x}_{\timeind-1}} \vec \mu_{\timeind} \right)\T
	+ 
	\vec \Sigma_x + \Sigma_f(\tilde{\vec \mu}_{\timeind}) \,, \\
	\cov(\vec x_{\timeind}, \vec \theta) 
	&\approx
	\cov(\vec x_{\timeind-1}, \vec \theta) 
	+ \nabla_{\vec \theta} \vec \mu_{\timeind} \cov(\vec x_{\timeind-1}, \vec \theta)\T 
	+ \nabla_{\vec \theta} \vec \mu_{\timeind} \vec \Sigma_{\theta} \,.
	\end{split}
\end{align*}
Note that $\nabla_{\vec x_{\timeind-1}} \vec \mu_{\timeind} = \mat I + \nabla_{\vec x_{\timeind-1}} \mu_f$ with the $\Delta$-transition model, and that $\nabla_{\tilde{\vec x}_{\timeind-1}} \vec \mu_{\timeind} \in \R^{\dimx \times (\dimx + \dimu + \dimp)}$. Derivatives of $\vec \mu_{\timeind}$ and $\vec \Sigma_{\timeind}$ with respect to $\vec \mu_{\timeind-1}$, $\vec \Sigma_{\timeind-1}$ and $\hat{\vec u}_{\timeind-1}$ are calculated following the standard rules of matrix calculus, and requires second-order derivative information of $f$ or the GP prediction.

It is common in GP regression to use zero-mean GP priors ($m_{(d)}(\cdot) \defeq 0$) to simplify calculations. The zero-mean prior is suitable for the $\Delta$-transition state space model formulation \citep{Ko2007, deisenroth2011_pilco}.

\section{\new{Design of Experiments Algorithm}}
\label{app:doe_algorithm}

\SetKwInOut{Input}{Input}
\SetKwInOut{Output}{Output}

\begin{algorithm}[H]
    \Input{
    \begin{minipage}[t]{10cm}%
     \strut
     Data points $\mat U$ (controls) and $\mat Y$ (measurements);

     Model candidates $\M_1, \dots, \M_M$;
     
     Feasible state spaces $\X_1, \dots, \X_M$ and control space $\U$;
     
     Number of training data points $N_1, \dots, N_M$ for GP surrogate models;
     
     Experiment budget $j_\text{max}$ (max no.\ of additional experiments);
     
     Model relevance threshold $\chi_{\text{lower}}^2$, such as $1\textsc{e-}3$;
     \strut
   \end{minipage}%
   }
    \Output{Selected model $\M^\ast$ with parameter estimate $\vec \theta^\ast \sim \N(\thetamap^\ast, \vec \Sigma_\theta^\ast)$;}
    \BlankLine
    \For{$j \leftarrow 1$ \KwTo $j_\text{max}$}{
        \For{$i \leftarrow 1$ \KwTo $M$}{
            \emph{Available data $\mat U$ and $\mat Y$ used for parameter estimation}\;
            $\vec \thetamap_i \leftarrow$ Model $\M_i$'s maximum \textit{a posteriori} parameter estimate\;
            \BlankLine
            \If{model $\M_i$ is a black box}{
                \emph{Construct GP surrogate model training data}\;
                % $\tilde{\mat X}_t, \tilde{\mat U}_t \leftarrow$ sample $N$ states and controls from $\X_i$ and $\U$\;
                $\tilde{\mat X}_t \leftarrow$ sample $N_i$ states from $\X_i$\;
                $\tilde{\mat U}_t \leftarrow$ sample $N_i$ controls from $\U$\;
                $\tilde{\vec \Theta}_t \leftarrow$ sample $N_i$ model parameters from $\N(\vec \thetamap_i, \epsilon \cdot \mat I)$\;
                Initialise $\tilde{\mat X}_{t+1} \in \R^{N_i \times \dimx}$ as empty array\;
                \For{$n \leftarrow 1$ \KwTo $N_i$}{
                    $\tilde{\mat X}_{t+1,n} \leftarrow f_i(\tilde{\mat X}_{t,n}, \tilde{\mat U}_{t,n}, \tilde{\vec \Theta}_{t,n})$\;
                }
                \BlankLine
                \emph{Learn GP surrogate model hyperparameters $\vec \lambda_i$}\;
                $\vec \lambda_i \leftarrow 
                    \argmax_{\vec \lambda} 
                    p\left(\tilde{\mat X}_{t+1}\,\middle|\, \tilde{\mat X}_{t}, \tilde{\mat U}_{t}, \tilde{\vec \Theta}_{t}, \vec \lambda \right)$
            }
            \BlankLine
            \emph{Available data $\mat U$ and $\mat Y$ used for parameter estimation}\;
            $\vec \Sigma_{\theta,i} \leftarrow$ Laplace approximation of model parameter covariance\;
            
            \BlankLine
            \emph{Available data $\mat U$ and $\mat Y$ used for model relevance check}\;
            $\chi_i^2 \leftarrow$ Model $\M_i$'s $\chi^2$ score\;
        }
        \BlankLine
        \emph{Select models with sufficiently high $\chi^2$ score}\;
        $\M_{\lbrace \rbrace} \leftarrow \lbrace \M_i \text{ if } \chi_i^2 \geq \chi_{\text{lower}}^2 \rbrace$\;
        \BlankLine
        \eIf{$\left| \M_{\lbrace \rbrace} \right| \leq 1$}{
            \emph{Best model found, or all models discarded}\;
            \textbf{BREAK}\;
        }{
            \emph{Generate additional experimental data}\;
            $\vec u_j^\ast \leftarrow$ solution to design of experiments problem \eqref{eq:design_problem_formulation} for models $\M_{\lbrace \rbrace}$\;
            $\vec y_j \leftarrow$ measurements from running experiment with controls $\vec u_j^\ast$\;
            % $\mat U, \mat Y \leftarrow \text{concatenate}(\mat U, \vec u_j^\ast), \,\text{concatenate}(\mat Y, \vec y_j)$\;
            $\mat U \leftarrow \text{concatenate}(\mat U, \vec u_j^\ast)$\;
            $\mat Y \leftarrow \text{concatenate}(\mat Y, \vec y_j)$\;
        }
    }
    \BlankLine
    \eIf{$\left| \M_{\lbrace \rbrace} \right| = 1$}{
        \emph{Success! One model $\M^\ast$ remaining}\;
    }{
        \emph{Inconclusive result! More than one model (or no models) remaining}\;
    }
    \caption{\new{The design of experiments process.}}
    \label{alg:doe}
\end{algorithm}

\section{Yeast Fermentation Case Study}
\label{app:yeast_fermentation_cs}

The yeast fermentation case study is taken from \cite{EspieMacchietto1989}, with constants \remove{e.g.}\new{(for example}~for true parameter values and noise covariances\new{)} taken from \cite{ChenAsprey2003}. There are $\dimx=2$ latent states (biomass and substrate concentration, respectively) and $\dimu=2$ control inputs (feed velocity and feed substrate concentration). We observe both states, hence $\dimz=\dimy=2$ and $\mat H_i = \mat H = \mat I$, with $\dimpi \in \lbrace 3, 4 \rbrace$ model parameters. For simplicity, we omit the model index and time step index when writing out the models
%Model $\M_1$ assumes Monod kinetics with constant specific death rate
\begin{align*}
    \M_1 \,:\,\,\left\lbrace
    \begin{aligned}
        \frac{\mathrm{d}x_1}{\mathrm{d}\time} &= (r - u_1 - \theta_4) x_1 \,, \\
        \frac{\mathrm{d}x_2}{\mathrm{d}\time} &= -\frac{rx_1}{\theta_3} + u_1(u_2-x_2) \,, \\
        r &= \frac{\theta_1 x_2}{\theta_2 + x_2} \,.
    \end{aligned}
    \right.
\end{align*}
%Model $\M_2$ assumes Contois kinetics with constant specific death rate
\begin{align*}
    \M_2 \,:\,\,\left\lbrace
    \begin{aligned}
        \frac{\mathrm{d}x_1}{\mathrm{d}\time} &= (r - u_1 - \theta_4) x_1 \,, \\
        \frac{\mathrm{d}x_2}{\mathrm{d}\time} &= -\frac{rx_1}{\theta_3} + u_1(u_2-x_2) \,, \\
        r &= \frac{\theta_1 x_2}{\theta_2 x_1 + x_2} \,.
    \end{aligned}
    \right.
\end{align*}
%Model $\M_3$ assumes linear specific growth rate
\begin{align*}
    \M_3 \,:\,\,\left\lbrace
    \begin{aligned}
        \frac{\mathrm{d}x_1}{\mathrm{d}\time} &= (r - u_1 - \theta_3) x_1 \,, \\
        \frac{\mathrm{d}x_2}{\mathrm{d}\time} &= -\frac{rx_1}{\theta_2} + u_1(u_2-x_2) \,, \\
        r &= \theta_1 x_2 \,.
    \end{aligned}
    \right.
\end{align*}
%Model $\M_4$ assumes Monod kinetics with constant maintenance energy
\begin{align*}
    \M_4 \,:\,\,\left\lbrace
    \begin{aligned}
        \frac{\mathrm{d}x_1}{\mathrm{d}\time} &= (r - u_1) x_1 \,, \\
        \frac{\mathrm{d}x_2}{\mathrm{d}\time} &= -\frac{rx_1}{\theta_3} + u_1(u_2-x_2) \,, \\
        r &= \frac{\theta_1 x_2}{\theta_2 + x_2} \,.
    \end{aligned}
    \right.
\end{align*}
Data is generated from $\M_1$ with parameters $\vec \theta = [0.25,\, 0.25,\, 0.88,\, 0.09]$ \citep{ChenAsprey2003}. The measurement noise covariance is assumed known and given by
\begin{align*}
    \vec \Sigma_y = \begin{bmatrix} 0.06 & -0.01 \\ -0.01 & 0.04 \end{bmatrix} \,,
\end{align*}
and there is no process noise, hence $\vec \Sigma_x \defeq \vec 0$. We start without any experimental data and initial model parameter estimates $\theta_{i,d} = 0.5$ and covariance $\vec \Sigma_{\theta,i} = 0.05 \cdot \mat I$ for all models $\M_i$ and $d \in \lbrace 1, \dots, \dimpi \rbrace$. The reason for this is the difficulty in finding initial experimental conditions that do not immediately render one or more models obviously inadequate. The initial states are given by $x_1 = 1$ and $x_2 = 0.01$, with initial latent state covariance $\vec \Sigma_0 = \diag(10^{-3}, 10^{-6})$. The controls have bounds $u_1 \in [0.05,\, 0.2]$ and $u_2 \in [5,\, 35]$. We let the control inputs have covariance given by $\vec \Sigma_u = \diag(10^{-6}, 10^{-3})$. We simulate 72 hours of fermentation, with measurements and changes in control signal every 1.5 hours.

\end{document}